\documentclass[11pt]{article}

\newcommand{\hqpool}{\textsc{DCLM-Baseline}\xspace}
\newcommand{\dclmrw}{\textsc{DCLM-RefinedWeb}\xspace}
\newcommand{\rawpool}{\textsc{DCLM-Pool}\xspace}

\usepackage[dvipsnames]{xcolor}
\usepackage[preprint]{acl}
\usepackage{ragged2e}
\usepackage{times}
\newcommand{\green}[1]{\textcolor{Green}{#1}}

\usepackage{latexsym}
\usepackage{colortbl}  %
\usepackage{makecell}
\usepackage[utf8]{inputenc} %
\usepackage[T1]{fontenc}    %
\usepackage{hyperref}       %
\usepackage{url}            %
\usepackage{booktabs}       %
\usepackage{amsfonts}       %
\usepackage{nicefrac}       %
\usepackage{microtype}      %
\usepackage{xspace}
\usepackage{multirow}
\usepackage{inconsolata}

\usepackage[capitalize]{cleveref}
\crefname{section}{Sec.}{Secs.}
\Crefname{section}{Section}{Sections}
\Crefname{table}{Table}{Tables}
\crefname{table}{Tab.}{Tabs.}
\Crefname{appendix}{Appendix}{Appendices}
\crefname{appendix}{Appx.}{Apps.}

\newcommand{\lowvar}{\textsc{Core\_V2}\xspace}

\newcommand{\strack}{\texttt{1B-1x}\xspace}

\newcommand{\ltrack}{\texttt{7B-1x}\xspace}
\newcommand{\xltrack}{\texttt{7B-2x}\xspace}
\newcommand{\justext}{\texttt{jusText}}
\newcommand{\resiliparse}{\texttt{resiliparse}}
\newcommand{\res}{\texttt{res.}}
\newcommand{\traf}{\texttt{traf.}}
\newcommand{\just}{\texttt{jusT.}}
\newcommand{\trafilatura}{\texttt{trafilatura}}
\newcommand{\fasttext}{\texttt{fastText}}

\usepackage[colorinlistoftodos]{todonotes}

\title{Beyond a Single Extractor: Re-thinking HTML-to-Text Extraction for LLM Pretraining}

\author{%
  Jeffrey Li$^{3*}$\And
  Josh Gardner$^{1\circ}$\And
  Doug Kang$^1$ \And
  Fangping Shi$^1$ \AND
  Karanjeet Singh$^1$ \And
  Chun-Liang Li$^1$ \And
  Herumb Shandilya$^2$ \And
  David Hall$^2$ \AND
  Oncel Tuzel$^1$ \And
  Percy Liang$^2$ \And
  Ludwig Schmidt$^2$ \And
  Hadi Pour Ansari$^1$ \And
  Fartash Faghri$^1$ \AND
  \vspace*{-20pt}\\
  $^1$Apple $^2$Stanford $^3$University of Washington \\
  \texttt{jwl2162@cs.washington.edu,fartash@apple.com}\\
  {\footnotesize
$^*$Work done during an internship at Apple.
$^{\circ}$Work done while at Apple.
}
}

\begin{document}

\maketitle

\begin{abstract}
    One of the first pre-processing steps for constructing web-scale LLM pretraining datasets
involves extracting text from HTML. Despite the immense diversity of web content, existing open-source datasets predominantly apply a single fixed extractor to all webpages. In this work, we investigate whether this practice leads to suboptimal coverage and utilization of Internet data.  We first show that while different extractors may lead to similar model performance on standard language understanding tasks, the pages surviving a fixed filtering pipeline can differ substantially. This suggests a simple intervention: by taking a \textit{Union} over different extractors, we can increase the token yield of \hqpool by up to 71\% while maintaining benchmark performance. We further show that for structured content such as tables and code blocks, extractor choice can significantly impact downstream task performance, with differences of up to 10 percentage points (p.p.) on WikiTQ and 3 p.p. on HumanEval.

\end{abstract}
\section{Introduction}

Large language models (LLMs) are primarily pretrained on webpages crawled from the Internet, such as from Common Crawl~\citep{commoncrawl}. A key early step for building such web-scale training datasets is to convert each page's HTML contents into plaintext. This is crucial not only because we wish to interact with models via natural language but also because HTML can contain a lot of unhelpful and auxiliary boilerplate such as navigation bars, hidden elements, and visual styling.

\begin{figure}
    \centering
    \includegraphics[width=0.95\linewidth]{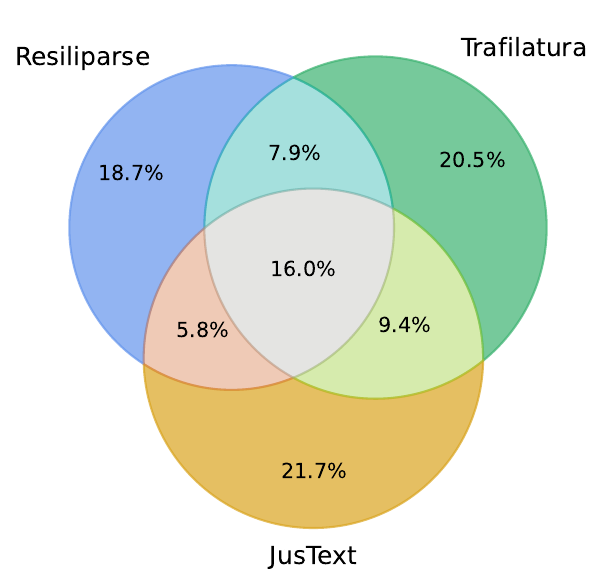}
    \caption{\textbf{Different extractors lead to different final pages.} The Venn diagram shows the overlap in resulting pages that come from applying the \hqpool pipeline to the outputs of three different initial extractors. 61\%  of pages are uniquely kept for just one extractor.}

    \label{fig:venn_diagram}
\end{figure}

Yet, while most existing research for LLM pretraining datasets has focused on filtering, cleaning, and deduplicating already extracted text, relatively little attention has been placed on extraction itself. Indeed, the standard practice for leading open-source datasets has been to select an existing rule-based extraction package and apply it uniformly to all of {Common Crawl}~\citep{commoncrawl}. For instance, \hqpool~\citep{dclm} uses \texttt{resiliparse}~\citep{resiliparse1, resiliparse2}, FinedWeb-Edu~\citep{lozhkov2024fineweb-edu} uses \texttt{trafilatura}~\citep{trafilatura}, and Nemotron-CC~\citep{su2024nemotroncc} uses \texttt{jusText}~\citep{justext1, justext2}.
These choices were motivated by qualitative inspection and findings that extractors achieved comparable downstream performance on standard benchmarks, differing mainly in other aspects such as runtime efficiency~\citep{dclm} or token yield~\citep{su2024nemotroncc}.

In this work, we investigate whether this practice of applying a single extractor to all webpages leads to sub-optimal usage of Common Crawl data. We demonstrate that while different extractors may result in similar aggregate performance on standard language understanding benchmarks, this similarity masks two critical issues: (1) extractors can capture complementary subsets of the web; (2) they exhibit larger differences in their ability to handle structured elements such as tables and code blocks.

First, we show that different extractors can lead to capturing surprisingly distinct portions of the web. When applying a fixed post-extraction curation pipeline from \hqpool, we find that \resiliparse{}, \trafilatura{}, and \justext{} yield substantially different sets of pages, with only 39\% surviving across more than one extractor (see \Cref{fig:venn_diagram}). This complementarity suggests a straightforward practical intervention: by taking the union of surviving pages across multiple extractors, we can increase the token yield from the \hqpool pipeline by 71\% (and 58\% when running further deduplication) while maintaining performance on standard benchmarks. Further, this increased yield translates to improved performance in data-constrained settings which simulate the practical finiteness of the Internet.

Second, we examine webpage elements that require specialized formatting—tables and code blocks—finding that extractor choice has substantial downstream impact. When isolating portions of Common Crawl containing such elements, model performance on relevant tasks depends heavily on the extractor used. For tables, a 7B model trained on \resiliparse{} outputs outperforms those coming from \justext{} and \trafilatura{} by 10.3 and 8.2 p.p. (a 3.2$\times$
improvement) respectively on WikiTableQuestions~\citep{pasupat-liang-2016-inferring}. For code, \justext{} significantly underperforms both alternatives, with degradations up to 3.6 p.p. on HumanEval~\citep{Chen2021EvaluatingLL}. 

Overall, our findings challenge current extraction practices when curating general web data for pretraining. While prior work has largely found popular extractors to be interchangeable when measured by common downstream model evaluations, we show that extraction remains an under-explored axis for improving pretraining datasets. Through more systematic evaluation of three widely-used extractors (\resiliparse{}, \trafilatura{}, and \justext{}) on Common Crawl, we provide actionable guidance for practitioners: using multiple extractors in parallel can substantially improve data yield, and extractor choice can significantly impact performance based on content type. Our analysis demonstrates that more thoughtful use of existing extraction tools can already yield meaningful improvements in dataset curation.

\section{Related work}

In this section, we discuss existing extractors, their usage in pretraining datasets, and filtering pipelines for tables and code blocks. We focus on the most relevant works here and defer an expanded discussion to  \Cref{app:extended_related_work}. 

\paragraph{Text extraction approaches.} 
Earlier approaches like \justext{}~\citep{justext1,justext2} work by segmenting HTML into blocks and then classifying each as boilerplate or main content using heuristics based on features such as length, link density, and stopword frequency. Introduced around the same time, \texttt{boilerpipe}~\citep{boilerpipe} similarly utilizes shallow text features but in combination  with learned decision trees. More recently, \trafilatura{}~\citep{trafilatura} emphasizes a more balanced approach between precision and recall. It uses rule-based heuristics that can fall back to other extractors like \justext{} when needed, but differs in design philosophy by capturing diverse content types (lists, tables, comments) rather than focusing narrowly on full English sentences (as \justext{} does). Meanwhile, \resiliparse{}~\citep{resiliparse1, resiliparse2} prioritizes computational efficiency, using simple tag and regex-based rules implemented in optimized C++ code. 
In contrast, another line of approaches develops deep learning models as extraction classifiers, including \texttt{web2text}~\citep{web2text}, \texttt{boilernet}~\citep{boilernet}, and \texttt{neuscraper}~\citep{xu2024cleaner}. While such efforts may be promising, we limit our investigation to rule-based extractors, which are easier to scale and have been used by the largest open datasets. 

\paragraph{Choosing text extractors for LLM pretraining.} Creators of the largest public pretraining datasets have  either used the extractions provided by Common Crawl (i.e., WET files) or existing rule-based extractors. Datasets such as C4~\citep{c4, c4_ai2}, RPJ~\citep{rpj, rpjv2}, and Dolma~\citep{soldaini2024dolma} used WET files. Meanwhile, The Pile~\citep{pile} and more recent state-of-the-art efforts run other extractors to remove more boilerplate. The FineWeb~\citep{lozhkov2024fineweb-edu} and  RefinedWeb~\citep{refinedweb} datasets use \trafilatura{}, while DCLM~\citep{dclm}  prefers \resiliparse{} due to its speed. In contrast, Nemotron-CC~\citep{su2024nemotroncc} uses \justext{} over \trafilatura{} as it yielded more tokens after running the FineWeb-Edu filtering classifier. Likely due to poorer scalability, neural-based extractors have been tested minimally for large-scale pretraining, though \citet{xu2024cleaner} show promising results for $\leq$ 410M parameter models. Notably, extractors were also traditionally evaluated against human extractions, but these metrics don't always predict how well LLMs can be trained on extractor outputs. For instance, \citet{bevendorffcomparison} finds \trafilatura{} to significantly outperform \justext{} but the two perform similarly in \citet{su2024nemotroncc}.

\paragraph{Filtering for tables and code blocks.} For tables, the most relevant work is The WDC Web Table Corpora~\citep{wdccommonstables, dresdentables}. This work filters \texttt{<table>} elements from two 2012 and 2015 Common Crawl dumps using various heuristics and classifies them based upon their purpose (e.g., whether the \texttt{<table>} contains true relational data  or exists just for layout purposes). 
For code blocks, Redstone-Code~\citep{redstone} finds \texttt{<code>} elements, filters for genuine code via regular expressions, and then extracts the parent pages using the same approach as WET files.
A more recent concurrent work is Nemotron-CC-Math~\citep{nemotronccmath}, which takes URLs from leading mathematics datasets and re-extracts them using the \texttt{lynx} browser in combination with Phi-4~\citep{abdin2024phi4technicalreport}. While they did not directly target code, they find that their dataset contains many pages with code snippets and improves their extractions, leading to better code generation capabilities. Overall, our work uses similar filtering methods for constructing datasets to study the role of extraction (rather than to compete with these approaches). We defer more detailed comparisons to \Cref{sec:tables,sec:codeblocks} as well as \Cref{app:extended_related_work}.

\section{Different extractors lead to different high-quality pages}\label{sec:extractor_union}

\begin{table*}[t!]
    \centering
    \small
    \caption{\textbf{Using multiple extractors increases token yield while maintaining performance.} We compare the token yields and \lowvar{} performance of using different extractors in the \hqpool curation pipeline. ``Union'' refers to combining the resulting pages from the individual extractors. The ``Random'' and ``Manual'' variants  refer to whether the pages in the intersection are selected randomly or according to a manually selected preference ordering: i.e., \resiliparse, then \trafilatura, then \justext. This is also the ordering of the percentiles for Union methods in the ``\fasttext{} Threshold'' column. Note that while we show MMLU performance at all scales for completeness, we advise caution when drawing conclusions from it at \strack{} and \ltrack{} scales due to potential noise. }
    \begin{tabular}{c|lcccc}
    \toprule
    Training Scale & Extractor & \fasttext{} Thresholds & Token Yield & \lowvar & MMLU \\
    \midrule
    \multirow{5}{*}{\parbox{3.0cm}{\centering \strack \\ (29B Tokens)}} & \resiliparse{}  & 0.11 & 39B & 28.5 & 24.7 \\
    & \trafilatura{}  & 0.11 & 27B & \textbf{29.6} & 25.7 \\
    & \justext{}      & 0.11 & 25B &  29.3 & \textbf{25.8}  \\
    & Union (Random)  & (0.11, 0.11, 0.11) & 55B & {28.1} & 25.1 \\
    & Union (Manual)  & (0.11, 0.11, 0.11) & \textbf{57B} & {29.2} & 24.1  \\ 
    \midrule
    \multirow{5}{*}{\parbox{3.0cm}{\centering \ltrack \\ (138B Tokens)}} & \resiliparse{}  & 0.11 & 193B & {42.6} & 39.5 \\
    & \trafilatura{}  & 0.11 & 135B & {41.9} & 37.7\\
    & \justext{}      & 0.11 & 124B & {41.2} & 34.5 \\
      & Union (Random)  & (0.11, 0.11, 0.11) & 273B & {41.3} & 33.6 \\
    & Union (Manual)  & (0.11, 0.11, 0.11) & \textbf{283B} & \textbf{43.1} & \textbf{43.5} \\ 
    \midrule
    \multirow{7}{*}{\parbox{3.0cm}{\centering \xltrack \\ (276B Tokens)}} & \resiliparse{}  &  0.11 & 386B  & {47.4} & \textbf{52.9} \\
    & \resiliparse{}  & 0.15  & 540B & {47.6} & 51.7 \\
    & \resiliparse{} & 0.18 & 650B  & {47.1} & 47.4 \\
    & Union (Random)  & (0.11, 0.11, 0.11) & 546B & {47.8} & 52.3 \\
    & Union (Manual)  & (0.11, 0.11, 0.11) & 565B & {47.9} & 51.4 \\
    & Union (Random)  & (0.11, 0.15, 0.15) & 639B  & \textbf{48.0} & 51.0 \\ 
    & Union (Manual)  & (0.11, 0.15, 0.15) & \textbf{662B} & {47.6} & 50.8  \\ 
    \bottomrule
\end{tabular}
    \label{tab:extractor_union}
\end{table*}

We first examine the impacts of extractor choice on standard English benchmarks, given a fixed (high-quality) post-extraction data pipeline.  

\subsection{Experiment setup} We start with the original Common Crawl WARCs that correspond to the \strack, \ltrack, and \xltrack \footnote{ As in DCLM, the scale names refer to the model size and the number of training tokens, given as a multiplier of the Chinchilla optimal amount~\citep{chinchilla}.} versions of \dclmrw~\citep{dclm} and re-extract these pages using three different extractors. While DCLM originally starts from \resiliparse{} extractions (with \texttt{main\_content} set to \texttt{True}), we now also try \trafilatura{} and \justext{} (both with default arguments) and then in parallel repeat \hqpool's filtering and deduplication procedures. 

Given these three different versions of \hqpool, we then assess the overlaps in page IDs\footnote{This is determined by the concatenation of the \texttt{WARC-Record-ID} and \texttt{WARC-Date} metadata keys.} when taking \textit{Unions} where we combine all surviving pages across versions and remove duplicate IDs (i.e., for pages where multiple extractions survive, we keep only one version). When deduplicating page IDs, we try two strategies for picking which version to keep: \textit{Random} chooses one surviving extraction arbitrarily while \textit{Manual} defines a hard-coded preference ordering favoring larger token yield: i.e., first \resiliparse{}, second \trafilatura{}, and third \justext{}.  

\begin{figure*}[t!]
    \centering
    \includegraphics[height=0.303\linewidth]{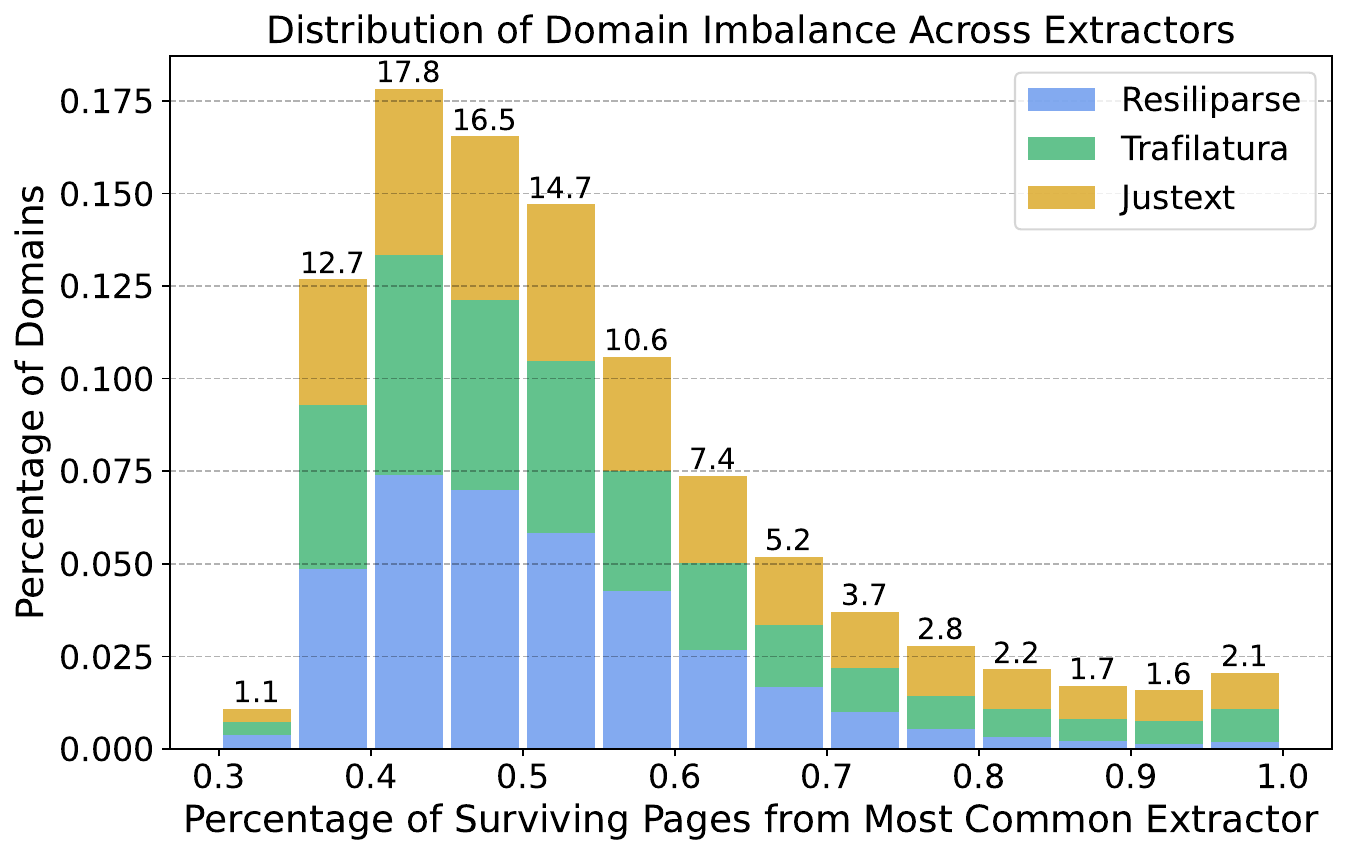}   
    \includegraphics[height=0.303\linewidth]
    {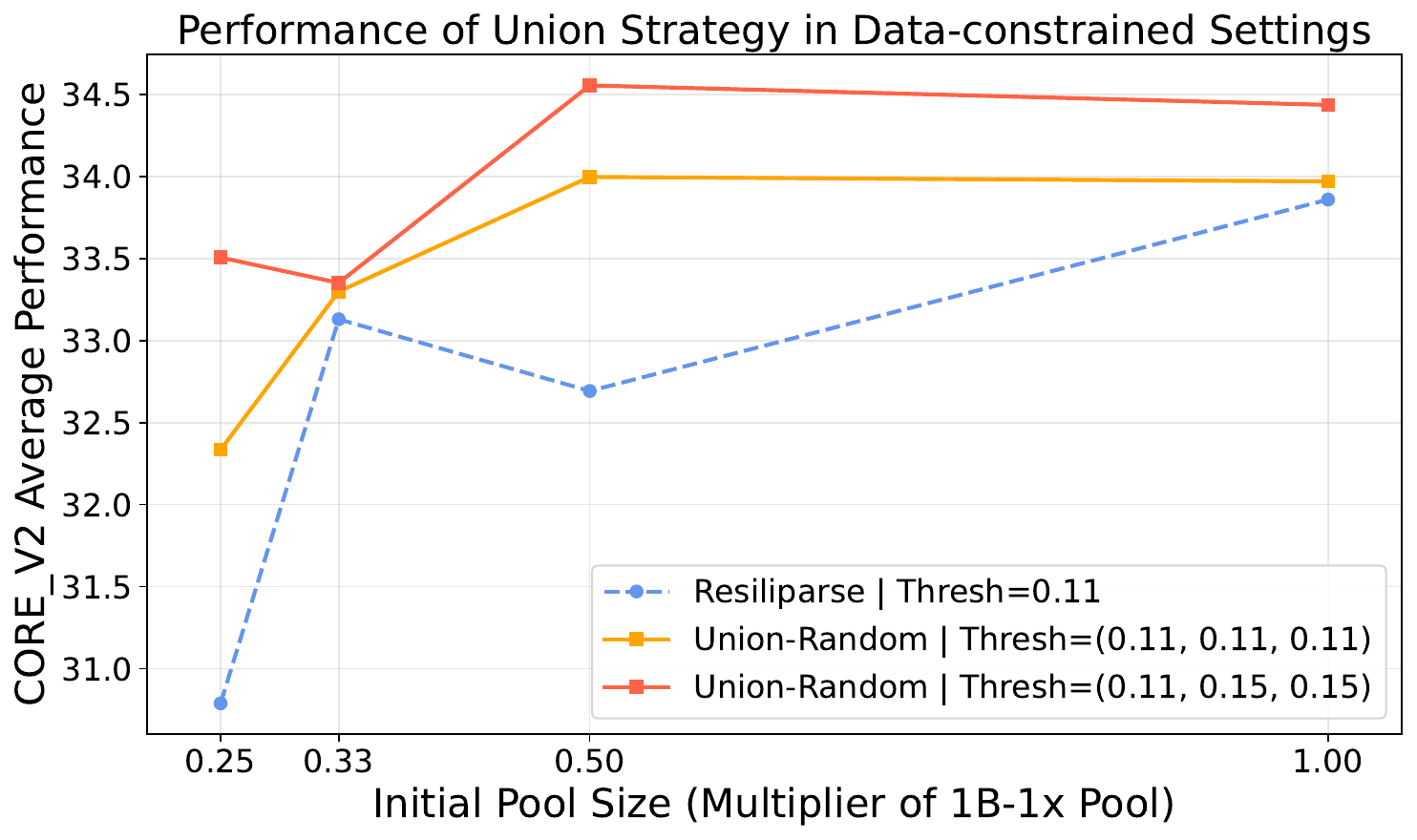}    
    \caption{\textbf{(Left) Analysis of extractor imbalance across domains.} We group pages from all three extractors (Top 11\%) by domain. For every domain with at least 50 pages, we compute the maximum ratio represented by any one extractor, plotting the distribution above. For 26.7\% and 7.6\% of domains respectively, at least 60\% and 80\% of surviving pages come from just one extractor. \textbf{(Right) Does higher yield lead to better performance when data-constrained?} We train \texttt{1B-5x} models on smaller subsamples of data curated from the \strack{} raw pool. The Union datasets that yield more tokens are able to achieve better performance. See \Cref{tab:union_data_constrained} for  extended results. }
    \label{fig:union_data_constrained}
\end{figure*}

\subsection{Results}

We start by comparing existing extractors in our setup. Then, we show that the resulting datasets contain many non-overlapping pages, allowing us to significantly improve data yield by combining them in our Unions. Finally, we examine the benefits of this approach in data-constrained settings.

\textbf{How do existing extractors perform on standard tasks?}  Across our \strack and \ltrack results in \Cref{tab:extractor_union}, we observe no consistent winner among extractors.  On \lowvar{}, \trafilatura{} and \justext{} perform similarly while \resiliparse{} shows lower a score at \strack{} but better performance at \ltrack{}. However, based on a preliminary study of run-to-run variation at \strack{} scale (see \Cref{tab:run_variation}), we find that these differences are likely within the noise of single runs—when averaged across three runs, the means (and standard deviations) of \resiliparse{} and \trafilatura{} become $28.3 \pm 0.7$ and $28.6 \pm 1.0$ respectively. Meanwhile, MMLU is close to random chance at \strack{} and more variable at \ltrack{} due to insufficient training. 

\textbf{Does combining extractors lead to better performance?}  As evidenced by the increases in token counts (\Cref{tab:extractor_union}) and overlap statistics (\Cref{fig:venn_diagram}) for Union datasets, different extractors yield substantially different pages. Further, these differences can be systematic: \Cref{fig:union_data_constrained} (left) shows that on a non-trivial portion of web domains, the vast majority of surviving pages come from a single extractor.
Overall, our Union datasets maintain performance on DCLM's \lowvar{} evaluations while increasing token yield by up to 71\% when we relax filtering thresholds for \trafilatura{} and \resiliparse{} to top-15\% (bringing their individual yields closer to \resiliparse{}'s). Meanwhile, we also see that performing a Union over high-quality pages from each extractor beats simply loosening the filtering threshold from the original \resiliparse{}-based dataset when targeting a similar token count. That is, we are able to \textit{recover} high-quality pages better when applying stricter filters across different extractors.

We note that our results do not prescribe an \textit{optimal number or set} of existing extractors to combine. We chose to study three approaches that are each battle-tested at scale but follow different heuristics and design choices. Adding more extractors may yield diminishing returns if they follow similar designs, or could perhaps significantly increase coverage if they are markedly different (e.g., neural-based or domain-specific). That said, our results remain a strong proof-of-concept that existing single-extractor approaches can be improved.

\textbf{Does the Union operation simply reintroduce duplicates?} Despite  deduplicating by page IDs, it is possible that the Union operation could re-introduce some fuzzy duplicates that were previously removed when running deduplication separately for each extractor. More concretely, suppose pages $x_1$ and $x_2$ are fuzzy duplicates both when we use extractors $e_1$ and $e_2$. It could be that the individual deduplication runs arbitrarily end up removing $e_1(x_1)$ and $e_2(x_2)$, leading to the Union containing a pair of fuzzy duplicates in $e_1(x_2)$ and $e_2(x_1)$. Whereas, if the same page was kept from both extractors, e.g., $e_1(x_1)$ and $e_2(x_1)$, the deduplication by page IDs would keep only one. To test whether this occurs, we run a second round of DCLM's fuzzy deduplication on our Union datasets and find that this decreases token count by 7-8\%. While not insignificant, the Union datasets still retain up to 58\% higher yield compared to \hqpool.

Meanwhile, we see from \Cref{tab:rededup} that rededuplication can have varying impacts on \lowvar{}. At \strack scale, the best result with rededuplication is 1.0 p.p. lower than the best result without it. Meanwhile, at \ltrack scale, rededuplication yields our best result of 44.1 which is 1.0 p.p. higher. It remains unclear whether these differences stem fully from variance or from more complex interactions between scale, union type, and rededuplication—for instance, whether systematic format differences in cross-extractor duplicates may benefit language modeling or whether Union (Random) tends to combine better with rededuplication.

\textbf{Does the Union operation lead to better performance under data constraints?} We explore training models in settings where data is more constrained by the initial amount of ``raw'' Internet data. This simulates the modern reality (albeit at a much smaller scale) that the largest frontier model runs may be approaching the limits of available Internet data. Specifically, we train \texttt{1B-5x} models on progressively smaller random subsamples of our \strack{} datasets. This means the Union datasets which yield more tokens will allow for fewer effective repetitions during training (see \Cref{tab:union_data_constrained}). As shown in \Cref{fig:union_data_constrained}, this translates to larger performance gains when data constraints are more severe.

\begin{table}[ht!]
    \centering
    \small
    \setlength{\tabcolsep}{4pt}
     \caption{\textbf{Effect of reduplicating Union datasets.} We compare the token yields and \lowvar{} performance of our Union datasets with and without reduplication. ``Rededup'' refers to running another round of deduplication on the Union dataset (now across the pages from different extractors). Here we always take a Union over the (0.11, 0.11, 0.11) thresholds. All other nomenclature follows from \Cref{tab:extractor_union}.}
    \begin{tabular}{c|lcc}
    \toprule
    Scale & Union Type & Token Yield & \lowvar \\
    \midrule
    \multirow{4}{*}{\strack{}} & Random  & 55B & 28.1  \\
    & Manual  & \textbf{57B} & \textbf{29.2}  \\ 
    & Random w/ Rededup  & 51B & 28.2  \\
    & Manual w/ Rededup & 53B & 27.4 \\ \midrule
     \multirow{4}{*}{\ltrack{}} & Random  & 273B &  41.3 \\
     & Manual  & \textbf{283B} & 43.1 \\
    & Random w/ Rededup  & 253B & \textbf{44.1}  \\ 
    & Manual w/ Rededup  & 265B & 41.8  \\ 
    \bottomrule
\end{tabular}
\label{tab:rededup}
\end{table}

\section{Extraction of tables}\label{sec:tables}

Given the proliferation of tabular data in many real-world applications, table understanding is an important skill for LLMs that has received growing attention~\citep{torrbenchmark, t4, vanbreugel2024}. However, such capabilities are not often tested by those proposing data curation techniques. Meanwhile, Common Crawl contains a large amount of HTML tables~\citep{tablib} that could presumably be used to learn such skills but which have also been under-emphasized during curation. In this section, we examine how different extractors behave on such pages and how these differences impact downstream models using the WikiTQ tabular understanding task~\citep{pasupat-liang-2016-inferring}.  %

\subsection{CC-Tables: Filtering Common Crawl for data tables.} We first filter down Common Crawl to a set of pages that contain \texttt{<table>} elements which are likely to be data tables, referring to the resulting set of pages as \textit{CC-Tables}. Inspired by the WDC Web Table Corpora~\citep{wdccommonstables}, we also use a combination of structural (e.g., based on row/column counts and row consistency) and model-based heuristics while simplifying the implementations of both (see \Cref{app:extended_related_work}). 
Our dataset and classifier model also differ in that they were: (1) built using all pre-2023 Common Crawl dumps as opposed to only two from 2012 and 2015; (2) we focus on extracting whole pages rather than just segments surrounding tables.

We found that our structural heuristics still left in many tables that were unlikely to be helpful, such as sizing charts for clothing products, landing pages for forums, and empty calendars. We train a \fasttext{} classifier with such tables as negative examples and tables from higher quality URLs (e.g., English Wikipedia and \texttt{.gov} domains) as positives. We then use this model to score all tables and filter out pages where no table is classified as positive. Notably, this filtering pipeline for CC-Tables does not yet \textit{apply any downstream extractor}, allowing us to then compare different extractions of these pages while limiting potential bias. More details are provided about CC-Tables in \Cref{app:table_filtering}.

\subsection{Qualitative extractor differences}  From our sample of pages provided in \Cref{app:tab_examples}, we observe \justext{} generally removes tables altogether (also noted by \citet{pile}), while \resiliparse{} and \trafilatura{} format them in different ways. The former uses white-space delimiting but can sometimes improperly merge together columns, while \trafilatura{} tries to convert tables into a markdown format but can fail to keep cell contents (see \Cref{fig:table_example_4}). Given these differences, a natural question is how this affects performance under different \textit{test-time} serializations. The ToRR benchmark~\citep{torr} allows us to study this question as it implements running evaluations over seven different serializations, including  both space-delimited (referred to as ``concat'') and markdown.

\begin{table}[t!]
    \centering      
    \small
    \caption{\textbf{\resiliparse{} is the most effective extractor for CC-Tables pages.} We compare the performance of \resiliparse{}, \trafilatura{}, and \justext{} on pages from our CC-Tables dataset. For each extracted version of CC-Tables, we mix it with \hqpool in a 20\%-80\% mix, train models, and evaluate on WikiTableQuestions (WikiTQ). ``No CC-Tables'' refers to training on only \hqpool. }
    \begin{tabular}{lcccc}
    \toprule
    CC-Tables Extraction & Model Scale  & WikiTQ-Avg.  \\
    \midrule
    No CC-Tables & \xltrack & 1.1 \\
    \midrule
    \resiliparse{} & \ltrack  & \textbf{11.9} \\
    \trafilatura{} & \ltrack & 3.7\\
    \justext{} & \ltrack  & 1.6 \\ \bottomrule
    \end{tabular}  
    \label{tab:extractor_tables}
\end{table}

\begin{figure*}[t!]
    \centering
    \includegraphics[width=0.87\linewidth]{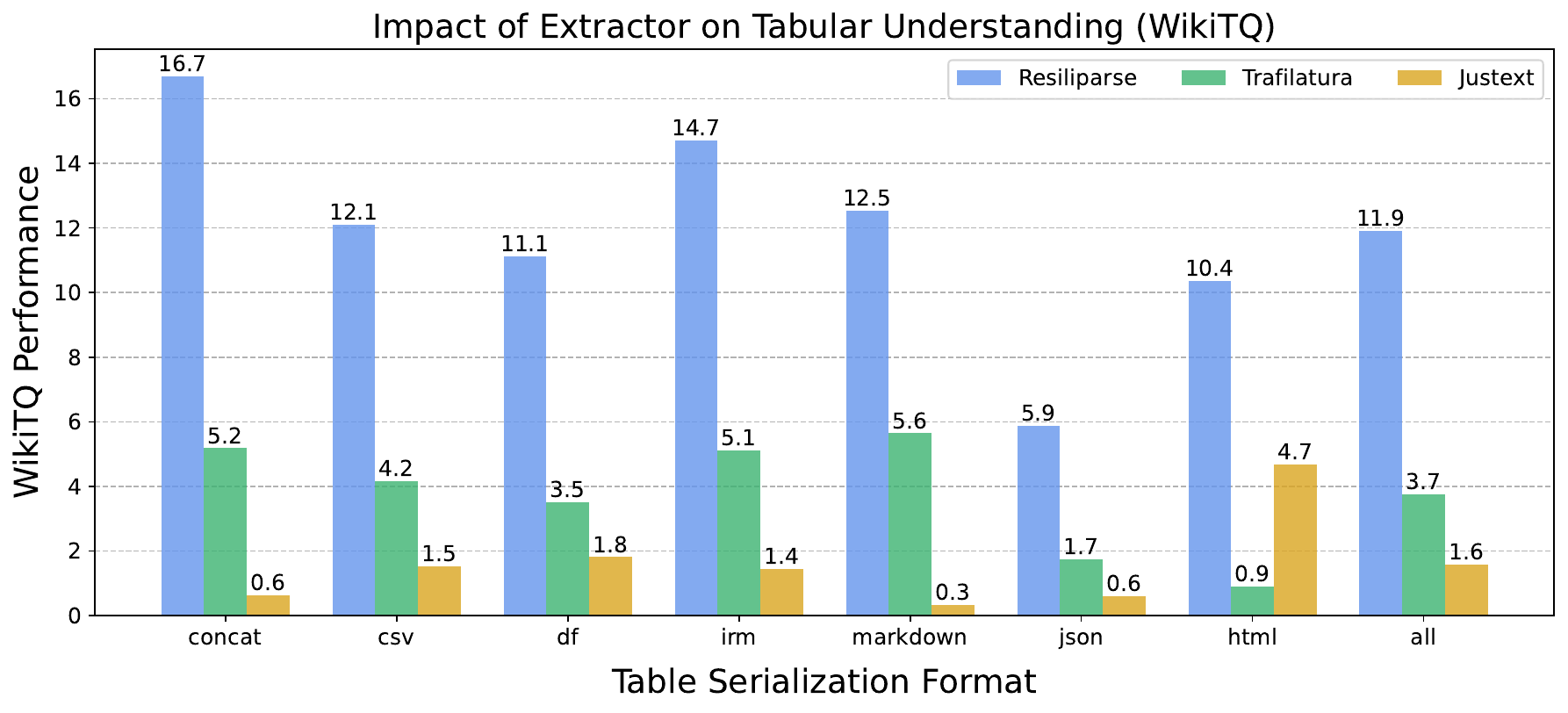}
    \caption{\textbf{Extractor performance remains consistent across serialization formats.} We plot the WikiTQ performance stratified across different test-time table serializations. Despite \resiliparse{} and \trafilatura{} producing tables that most closely match ``concat'' and ``markdown'' formats respectively, we observe a surprising degree of generalization across most serializations (with the exception of ``json'').   }
    \label{fig:wiki_tq_plot}
\end{figure*}

\begin{table*}[t!]
    \centering
    \small
    \setlength{\tabcolsep}{4pt}
    \caption{\textbf{CC-Tables-\resiliparse{} closes 72\% of the gap between the original DCLM-7B-8k model and Llama-3-8B.} The top rows show results for existing models while the bottom rows are new models obtained by replacing the original DD-style cooldown for DCLM-7B-8K with one that mixes CC-Tables-\resiliparse{} and \hqpool and performs WiSE-FT~\citep{wiseft}. We try different interpolation weights $\alpha \in [0.8, 1.0]$ with $\alpha=0.85$ providing a good balance between tabular and standard evaluations. }
    \begin{tabular}{lccccc}
    \toprule
    Model / Cooldown Dataset & Model Size &  Tokens  & WikiTQ-Avg. & \lowvar{} & MMLU \\
    \midrule

    LLama-3-8B & 8B & 15T  & \textbf{29.3} & 56.5 & 66.2 \\ 
    DCLM-7B-8k & 7B & 2.8T  & 5.0 & 56.0 & 63.7 \\ \midrule
    DCLM-Bas. + CC-Tables ($\alpha$ = 1.00) & 7B & 2.7T & \textbf{22.7} & 53.7 & 62.3 \\
    DCLM-Bas. + CC-Tables ($\alpha$ = 0.90) & 7B & 2.7T & 22.8 & 54.7 & 62.6 \\
    DCLM-Bas. + CC-Tables ($\alpha$ = 0.85) & 7B & 2.7T & 22.4 & 55.7 & 62.7 \\
    DCLM-Bas. + CC-Tables ($\alpha$ = 0.80)& 7B & 2.7T & 18.8 & \textbf{56.1} & \textbf{63.0} \\

    \bottomrule
    \end{tabular}  
    \label{tab:extractor_tables_cooldown}
\end{table*}

\subsection{Empirical Results}

\textbf{Training setup.} We train various 8K-context models on mixes of 80\% general web data (i.e., DCLM-Baseline) and 20\% CC-Tables after applying different extraction algorithms. To train longer context models efficiently, we use the Dataset Decomposition technique from~\citep{pouransari2024dataset}. The $4\times$ increase in context length compared to standard DCLM models is important for our evaluations as some serializations (e.g. HTML and JSON) in ToRR lead to examples exceeding a context length of 2048. For evaluation, we use a modified version of ToRR's~\citep{torr} implementation of WikiTableQuestions (WikiTQ)~\citep{pasupat-liang-2016-inferring}.  Specifically, we modify the set of post-processors that ToRR uses to avoid inflating the scores of some shorter generations (see \Cref{app:table_eval} for more details).

\textbf{Which extractors are best for CC-Tables?} From our results in \Cref{tab:extractor_tables}, we see that the qualitative differences between extractors indeed lead to significant quantitative differences on WikiTQ. Unsurprisingly, \justext{} performs the worst, while both \resiliparse{} and \trafilatura{} lead to non-trivial improvements over training on more tokens of DCLM-Baseline alone. Interestingly, \resiliparse{} performs markedly better than \trafilatura{}. We see from \Cref{fig:wiki_tq_plot} that while \resiliparse{} produces only space-delimited tables and favors ``Concat'' at test-time, it still leads to the best performance on all test formats. Meanwhile, \trafilatura{} leads to favoring markdown, but suffers worse performance overall. 

These results perhaps echo a similar finding from  \citet{llama3} that markdown markers can be harmful to models mostly trained on web-crawled data. We also observe from \Cref{tab:tables_appendix} in \Cref{app:tab_additional_results} that our results hold even if we modify the mixture weights and use the \trafilatura{} version of DCLM-Baseline (i.e., it is not the result of conflicting extractors between the different mixture components). However, we cannot rule out the role of other extraction differences besides formatting (e.g., how surrounding text is captured or whether all data cells are actually retained). 

\textbf{Can CC-Tables patch tabular performance for DCLM-7B-8k?} Finally, we see in \Cref{tab:extractor_tables_cooldown} whether CC-Tables-\resiliparse{} can help close the sizable gap between DCLM-7B-8k and LLama-3-8B on WikiTQ. While the original DCLM paper shows that the two models are comparable on standard language evaluations, we see here the latter is far better at tabular understanding. To bridge this gap, we find that replacing the final DD cooldown of DCLM-7B-8k with a 80\%-20\% mix of DCLM-Baseline and CC-Tables-\resiliparse{} can close 73\% of this gap. While the improvement on WikiTQ initially comes at the cost of standard metrics, a simple WiSE-FT~\citep{wiseft}  weight interpolation can help mitigate this, with the weight $\alpha=0.85$ providing a reasonable balance.

\begin{table*}[t!]
    \centering      
    \small
    \caption{\textbf{CC-Code extraction and code generation performance.} We observe that using \justext{} leads to the additional CC-Code pages harming HumanEval performance. Meanwhile, \resiliparse{} and \trafilatura{} perform comparably. In each mix we train on 25\% StackV2, 25\% CC-Code (when present), and fill in the remainder with \hqpool. The top rows are for cooldown runs on top of the DCLM-7B~\citep{dclm} checkpoint prior to their original cooldown at 2T tokens. The bottom rows correspond to training \xltrack models from scratch. All code generation metrics are reported as Pass@1 computed from 20 samples per problem.}
    \begin{tabular}{lccccc|c|cc}
    \toprule
    Dataset  & Python & Java & JS & C++ & Go & HE-Avg. & \lowvar{} & MMLU \\
    \midrule
    \multicolumn{9}{c}{\textit{Alternative 200B token cooldown for DCLM-7B}} \\
    \midrule
    \rowcolor{white}            
    \makecell[l]{\hqpool (Top-7\%) \\ \quad + Stack v2 \\ \quad + CC-Code-\res} & 26.7 & \textbf{29.9} & 24.0 & \textbf{26.6} & \textbf{17.7} & \textbf{25.0} &  54.7 & 60.8 \\
    \rowcolor{gray!10}
    \makecell[l]{\hqpool (Top-7\%) \\ \quad + Stack v2 \\ \quad + CC-Code-\traf} & 28.0 & 27.0 & \textbf{25.3} & 23.0 & 16.9 & 24.0 & 54.3 & 60.5 \\
    \rowcolor{white}
    \makecell[l]{\hqpool (Top-7\%) \\ \quad +  Stack v2 \\ \quad + CC-Code-\just} & 25.0 & 24.1 & 21.7 & 20.3 & 15.8 & 21.4 &  54.0 & 58.9 \\
    \rowcolor{gray!10}
    \makecell[l]{\hqpool (Top-7\%) \\ \quad + Stack v2} & \textbf{28.1} & 23.6 & 24.6 & 20.9 & 15.2 & 22.5 & \textbf{ 55.4} & \textbf{62.4} \\ 
    \midrule
    \multicolumn{9}{c}{\textit{Trained from scratch, DCLM \xltrack scale}} \\
    \midrule
    \rowcolor{white} 
    \makecell[l]{\hqpool (Original) \\ \quad + Stack v2 \\ \quad + CC-Code-\res} & 20.5 & 20.1 & 18.5  & \textbf{19.8} & 11.7 & 18.1 & 44.0 & 41.8 \\
    \rowcolor{gray!10}
    \makecell[l]{\hqpool (Original) \\ \quad + Stack v2 \\ \quad + CC-Code-\traf} & \textbf{23.2} & 20.8 & 18.6 & 16.6 & \textbf{13.8} & 18.6 & 43.0 & 41.1 \\
    \rowcolor{white}
    \makecell[l]{\hqpool (Original) \\ \quad + Stack v2 \\ \quad + CC-Code-\just} & 19.4 & 16.4 & 16.6 & 16.4 & 11.6 & 16.1 & 40.4 & 33.2  \\
    \rowcolor{gray!10}
    \makecell[l]{\hqpool (Original) \\ \quad + Stack v2} & 21.9 & \textbf{20.8} & \textbf{19.6  } & 19.5 & 13.0 & \textbf{18.9} & \textbf{47.4} & \textbf{50.7} \\
    \bottomrule
    \end{tabular}  
    \label{tab:code_results}
\end{table*}

\section{Extraction of code blocks}\label{sec:codeblocks}

We conduct similar experiments for code blocks as we do for tables, isolating pages that likely contain genuine code blocks and then exploring the impacts of different extractors on downstream models. Here, we evaluate on HumanEval~\citep{Chen2021EvaluatingLL} using the BigCode evaluation harness~\citep{bigcode-evaluation-harness} to measure the ability of models to write code completions based on provided function signatures and docstrings.

\subsection{CC-Code: filtering Common Crawl for code blocks} Similar to Redstone-Code~\citep{redstone}, we aim to identify genuine code blocks that are interleaved with natural text from the general web (e.g., tutorials, documentation, and forums) in an effort to complement traditional ``source-code'' datasets (e.g., created from GitHub) that are commonly used during pretraining. Our approach differs from that of \citet{redstone} in that we (1) focus on directly classifying \texttt{<pre>} elements instead of relying on the presence of child \texttt{<code>} elements; (2) train a model for said classification as opposed to using regular expressions. We do not claim nor evaluate whether our approach is better for curating a more performant dataset. Rather, our goal is to explore the impact of different extractors given a reasonable filtered set of code-related pages.

We limit our exploration to \texttt{<pre>} HTML elements, as the \texttt{<pre>} tag is commonly used to maintain pre-formatted text (i.e., preserving spacing such as tabs and line breaks). However, based upon qualitative inspection, we found that \texttt{<pre>} elements are also used often for other purposes such as to represent poems, lyrics, and emails. As such, we train a \fasttext{} classifier to filter out non-code usages of \texttt{<pre>}. We construct the training set for this classifier primarily by manually inspecting and labeling the 160 most frequent domains containing \texttt{<pre>} as either code or non-code related (ignoring a domain if ambiguous). After applying our classifier to score \texttt{<pre>} elements, we retain pages that contain at least one that is high-scoring. We refer to the surviving set of pages as CC-Code and then evaluate different extractors by applying them to CC-Code, before also using DCLM's English and \fasttext{} filters to improve overall quality. More details about our filtering pipeline for CC-Code are provided in \Cref{app:cc_code_filtering}.

\subsection{Qualitative extractor differences}  We present some examples from CC-Code in \Cref{app:code_examples}. Likely due to its dependence on stop-words, we find that \justext{} often ends up excluding code blocks. Meanwhile, we find that \trafilatura{} and \resiliparse{} both keep some code blocks but the former fails to preserve proper formatting (e.g., collapsing newlines and removing indentation). It is a natural question whether or not these misformattings might hinder code generation ability, or if additionally including traditional source-code (which is assumed to be properly formatted) in training is enough alleviate these issues.

\subsection{Empirical results}

\textbf{Training setup.} We train models in two different setups. First, we pretrain models from scratch using the hyperparamters from the DCLM \xltrack track (\Cref{tab:code_results}, bottom). Second, we start from DCLM-7B's pre-cooldown checkpoint (trained on 2T tokens) and modify the original 200B cooldown dataset (\Cref{tab:code_results}, top). Originally in DCLM, the cooldown dataset was a mix of more strictly filtered \hqpool (i.e., Top-7\%) mixed with 30\% math data. Here, since we care about code performance, we train on mixes that are at least 25\% code from Stack v2~\citep{starcoder2instruct} and optionally 25\% CC-Code. The remainder is then made up of \hqpool.

\textbf{What extractors are best for CC-Code?} As expected from our inspection, we observe from \Cref{tab:code_results} that \justext{} leads to sub-optimal extraction of CC-Code pages. In the ``cooldown'' setup, it performs 2.6 p.p. worse than \resiliparse{} and \trafilatura{}, and even 1.0 p.p. worse than replacing CC-Code with general web documents. Meanwhile, better utilization of CC-Code pages can be achieved with \trafilatura{} and \resiliparse{}, which are able to improve code performance, particularly for Java and C++. Between the two, \resiliparse{} offers slightly better performance, perhaps suggesting that its improved whitespace preservation helps languages where line breaks and indentation matter (i.e., JavaScript is a notable exception as it is more robust to whitespace removals). However, we do note that the absolute performance metrics are still close overall.

In the ``from-scratch'' setup,  \justext{} still leads to the worst performance but CC-Code is not as useful overall as dropping CC-Code altogether is slightly better than including it. We speculate that this may be due to source-code being more relevant to HumanEval-style evaluations than interleaved code, as the prompts are already in the form of partially written code. Therefore, when starting from scratch, it may be more important to build up a base of knowledge from seeing more source-code before interleaved data becomes additionally useful. In the ``cooldown'' setting, the checkpoint we start from has already been trained on a mix containing Stack v2, perhaps allowing for better usage of CC-Code. For both setups, we observe that (perhaps as expected), replacing \hqpool tokens with CC-Code tokens decreases standard evaluation performance, with bigger drops in the from-scratch setting.

\section{Conclusions}

We demonstrate that text extraction is an underexplored aspect of LLM pretraining data curation. While standard practice applies a single extractor to all of Common Crawl, we show this leads to suboptimal data yield—using multiple extractors in parallel can significantly increase yield while maintaining performance. We further show that extractor choice can significantly impact model performance depending on content type, with larger differences on tables and code tasks compared to standard language evaluations. We hope this work encourages practitioners to revisit extraction as a first-class component of their curation pipelines.

\section*{Limitations}

While we systematically study three popular extractors, we do not develop new extraction methods or sophisticated strategies for selecting between them based on page characteristics. Relatedly, these libraries remain under active development, so newer releases may patch specific failure modes and yield better results. We believe these are all promising directions for future work. Another limitation is that we did not extensively explore alternative filtering strategies for CC-Tables and CC-Code. Our primary aim was to develop reasonable pipelines for isolating pages with genuine tables and code blocks to study the impact of extractor choice. While we believe our conclusions about extractor performance on structured content should generalize across filtering approaches, improved filtering could yield higher absolute scores on downstream evaluations, such as those obtained by Nemotron-CC-Math \cite{nemotronccmath}. Finally, due to computational constraints, we were unable to fully assess run-to-run variation, particularly for our 7B results.

\section*{Ethical considerations}

Better extraction can lead to more effective coverage of Internet data, which can come with both benefits and risks. Exposing models to more diverse content could reduce potential blind spots in their capabilities. However, better extraction may also increase exposure to harmful, toxic, or inappropriate content. For instance, it may inadvertently increase the capture of personal information or copyrighted material that a given single extractor might miss. While we do not release datasets or models from our experiments, we recommend that practitioners using these techniques continue to implement robust content filtering and respect data usage rights. Finally, we note that larger datasets enable longer training runs with greater computational and environmental costs that should be weighed against their benefits.

\section*{Acknowledgements}

We would like to thank Rick Chang, Cem Koc, Stephen Pulman, Denise Hui, Elle Barnes, Merhawie Woldezion, and Pranay Pattani for valuable feedback, guidance, and support. We'd also like to thank Shir Ashury-Tahan, Yifan Mai, and Elron Bandel for helpful discussions related to setting up and using ToRR.

\clearpage

\bibliography{main}

\begin{thebibliography}{49}
\providecommand{\natexlab}[1]{#1}

\bibitem[{Abdin et~al.(2024)Abdin, Aneja, Behl, Bubeck, Eldan, Gunasekar, Harrison, Hewett, Javaheripi, Kauffmann, Lee, Lee, Li, Liu, Mendes, Nguyen, Price, de~Rosa, Saarikivi, Salim, Shah, Wang, Ward, Wu, Yu, Zhang, and Zhang}]{abdin2024phi4technicalreport}
Marah Abdin, Jyoti Aneja, Harkirat Behl, Sébastien Bubeck, Ronen Eldan, Suriya Gunasekar, Michael Harrison, Russell~J. Hewett, Mojan Javaheripi, Piero Kauffmann, James~R. Lee, Yin~Tat Lee, Yuanzhi Li, Weishung Liu, Caio C.~T. Mendes, Anh Nguyen, Eric Price, Gustavo de~Rosa, Olli Saarikivi, and 8 others. 2024.
\newblock \href {https://arxiv.org/abs/2412.08905} {Phi-4 technical report}.
\newblock \emph{Preprint}, arXiv:2412.08905.

\bibitem[{Allal et~al.(2025)Allal, Lozhkov, Bakouch, Blázquez, Penedo, Tunstall, Marafioti, Kydlíček, Lajarín, Srivastav, Lochner, Fahlgren, Nguyen, Fourrier, Burtenshaw, Larcher, Zhao, Zakka, Morlon, Raffel, von Werra, and Wolf}]{smollm2}
Loubna~Ben Allal, Anton Lozhkov, Elie Bakouch, Gabriel~Martín Blázquez, Guilherme Penedo, Lewis Tunstall, Andrés Marafioti, Hynek Kydlíček, Agustín~Piqueres Lajarín, Vaibhav Srivastav, Joshua Lochner, Caleb Fahlgren, Xuan-Son Nguyen, Clémentine Fourrier, Ben Burtenshaw, Hugo Larcher, Haojun Zhao, Cyril Zakka, Mathieu Morlon, and 3 others. 2025.
\newblock Smollm2: When smol goes big -- data-centric training of a small language model.
\newblock \emph{arXiv preprint arXiv:2502.02737}.

\bibitem[{Ashury-Tahan et~al.(2025{\natexlab{a}})Ashury-Tahan, Mai, C, Gera, Perlitz, Yehudai, Bandel, Choshen, Shnarch, Liang, and Shmueli-Scheuer}]{torrbenchmark}
Shir Ashury-Tahan, Yifan Mai, Rajmohan C, Ariel Gera, Yotam Perlitz, Asaf Yehudai, Elron Bandel, Leshem Choshen, Eyal Shnarch, Percy Liang, and Michal Shmueli-Scheuer. 2025{\natexlab{a}}.
\newblock The mighty torr: A benchmark for table reasoning and robustness.
\newblock \emph{arXiv preprint arXiv:2502.19412}.

\bibitem[{Ashury-Tahan et~al.(2025{\natexlab{b}})Ashury-Tahan, Mai, C, Gera, Perlitz, Yehudai, Bandel, Choshen, Shnarch, Liang, and Shmueli-Scheuer}]{torr}
Shir Ashury-Tahan, Yifan Mai, Rajmohan C, Ariel Gera, Yotam Perlitz, Asaf Yehudai, Elron Bandel, Leshem Choshen, Eyal Shnarch, Percy Liang, and Michal Shmueli-Scheuer. 2025{\natexlab{b}}.
\newblock The mighty torr: A benchmark for table reasoning and robustness.
\newblock \emph{arXiv preprint arXiv:2502.19412}.

\bibitem[{Bandel et~al.(2024)Bandel, Perlitz, Venezian, Friedman, Arviv, Orbach, Don-Yehiya, Sheinwald, Gera, Choshen, Shmueli-Scheuer, and Katz}]{bandel-etal-2024-unitxt}
Elron Bandel, Yotam Perlitz, Elad Venezian, Roni Friedman, Ofir Arviv, Matan Orbach, Shachar Don-Yehiya, Dafna Sheinwald, Ariel Gera, Leshem Choshen, Michal Shmueli-Scheuer, and Yoav Katz. 2024.
\newblock \href {https://aclanthology.org/2024.naacl-demo.21} {Unitxt: Flexible, shareable and reusable data preparation and evaluation for generative {AI}}.
\newblock In \emph{Proceedings of the 2024 Conference of the North American Chapter of the Association for Computational Linguistics: Human Language Technologies (Volume 3: System Demonstrations)}, pages 207--215, Mexico City, Mexico. Association for Computational Linguistics.

\bibitem[{Barbaresi(2021)}]{trafilatura}
Adrien Barbaresi. 2021.
\newblock \href {https://doi.org/10.18653/v1/2021.acl-demo.15} {Trafilatura: {A} web scraping library and command-line tool for text discovery and extraction}.
\newblock In \emph{Proceedings of the 59th Annual Meeting of the Association for Computational Linguistics and the 11th International Joint Conference on Natural Language Processing: System Demonstrations}, pages 122--131, Online. Association for Computational Linguistics.

\bibitem[{Ben~Allal et~al.(2022)Ben~Allal, Muennighoff, Kumar~Umapathi, Lipkin, and von Werra}]{bigcode-evaluation-harness}
Loubna Ben~Allal, Niklas Muennighoff, Logesh Kumar~Umapathi, Ben Lipkin, and Leandro von Werra. 2022.
\newblock A framework for the evaluation of code generation models.
\newblock \url{https://github.com/bigcode-project/bigcode-evaluation-harness}.

\bibitem[{Bevendorff et~al.(2023)Bevendorff, Gupta, Kiesel, and Stein}]{bevendorffcomparison}
Janek Bevendorff, Sanket Gupta, Johannes Kiesel, and Benno Stein. 2023.
\newblock \href {https://doi.org/10.1145/3539618.3591920} {An empirical comparison of web content extraction algorithms}.
\newblock In \emph{Proceedings of the 46th International ACM SIGIR Conference on Research and Development in Information Retrieval}, SIGIR '23, page 2594–2603, New York, NY, USA. Association for Computing Machinery.

\bibitem[{Bevendorff et~al.(2021)Bevendorff, Potthast, and Stein}]{resiliparse2}
Janek Bevendorff, Martin Potthast, and Benno Stein. 2021.
\newblock {FastWARC: Optimizing Large-Scale Web Archive Analytics}.
\newblock In \emph{International Symposium on Open Search Technology (OSSYM)}.
\newblock \url{https://github.com/chatnoir-eu/chatnoir-resiliparse}.

\bibitem[{Bevendorff et~al.(2018)Bevendorff, Stein, Hagen, and Potthast}]{resiliparse1}
Janek Bevendorff, Benno Stein, Matthias Hagen, and Martin Potthast. 2018.
\newblock {Elastic ChatNoir: Search Engine for the ClueWeb and the Common Crawl}.
\newblock In \emph{European Conference on Information Retrieval Research (ECIR)}.
\newblock \url{https://github.com/chatnoir-eu/chatnoir-resiliparse}.

\bibitem[{Chang et~al.(2024)Chang, Cui, Dong, Huang, Huang, Huang, Li, Lv, Ma, Sun et~al.}]{redstone}
Yaoyao Chang, Lei Cui, Li~Dong, Shaohan Huang, Yangyu Huang, Yupan Huang, Scarlett Li, Tengchao Lv, Shuming Ma, Qinzheng Sun, and 1 others. 2024.
\newblock {RedStone}: {Curating} general, code, math, and {QA} data for large language models.
\newblock \emph{arXiv preprint arXiv:2412.03398}.

\bibitem[{Chen et~al.(2021)Chen, Tworek, Jun, Yuan, Ponde, Kaplan, Edwards, Burda, Joseph, Brockman, Ray, Puri, Krueger, Petrov, Khlaaf, Sastry, Mishkin, Chan, Gray, Ryder, Pavlov, Power, Kaiser, Bavarian, Winter, Tillet, Such, Cummings, Plappert, Chantzis, Barnes, Herbert-Voss, Guss, Nichol, Babuschkin, Balaji, Jain, Carr, Leike, Achiam, Misra, Morikawa, Radford, Knight, Brundage, Murati, Mayer, Welinder, McGrew, Amodei, McCandlish, Sutskever, and Zaremba}]{Chen2021EvaluatingLL}
Mark Chen, Jerry Tworek, Heewoo Jun, Qiming Yuan, Henrique Ponde, Jared Kaplan, Harrison Edwards, Yura Burda, Nicholas Joseph, Greg Brockman, Alex Ray, Raul Puri, Gretchen Krueger, Michael Petrov, Heidy Khlaaf, Girish Sastry, Pamela Mishkin, Brooke Chan, Scott Gray, and 34 others. 2021.
\newblock \href {https://arxiv.org/abs/2107.03374} {Evaluating large language models trained on code}.
\newblock \emph{ArXiv preprint}, abs/2107.03374.

\bibitem[{{Common Crawl}({2007})}]{commoncrawl}
{Common Crawl}. {2007}.
\newblock {Common Crawl}.
\newblock \url{https://commoncrawl.org}.

\bibitem[{Computer(2023)}]{rpjv2}
Together Computer. 2023.
\newblock \href {https://github.com/togethercomputer/RedPajama-Data} {Redpajama: an open dataset for training large language models}.

\bibitem[{Dodge et~al.(2021)Dodge, Sap, Marasovi{\'c}, Agnew, Ilharco, Groeneveld, Mitchell, and Gardner}]{c4_ai2}
Jesse Dodge, Maarten Sap, Ana Marasovi{\'c}, William Agnew, Gabriel Ilharco, Dirk Groeneveld, Margaret Mitchell, and Matt Gardner. 2021.
\newblock \href {https://doi.org/10.18653/v1/2021.emnlp-main.98} {Documenting large webtext corpora: A case study on the colossal clean crawled corpus}.
\newblock In \emph{Proceedings of the 2021 Conference on Empirical Methods in Natural Language Processing}, pages 1286--1305, Online and Punta Cana, Dominican Republic. Association for Computational Linguistics.

\bibitem[{Eberius et~al.(2015)Eberius, Braunschweig, Hentsch, Thiele, Ahmadov, and Lehner}]{dresdentables}
Julian Eberius, Katrin Braunschweig, Markus Hentsch, Maik Thiele, Ahmad Ahmadov, and Wolfgang Lehner. 2015.
\newblock \href {https://doi.org/10.1109/BDC.2015.30} {Building the dresden web table corpus: A classification approach}.
\newblock In \emph{2015 IEEE/ACM 2nd International Symposium on Big Data Computing (BDC)}, pages 41--50.

\bibitem[{Eggert et~al.(2023)Eggert, Huo, Biven, and Waugh}]{tablib}
Gus Eggert, Kevin Huo, Mike Biven, and Justin Waugh. 2023.
\newblock Tablib: A dataset of 627m tables with context.
\newblock \emph{arXiv preprint arXiv:2310.07875}.

\bibitem[{Gao et~al.(2021)Gao, Biderman, Black, Golding, Hoppe, Foster, Phang, He, Thite, Nabeshima, Presser, and Leahy}]{pile}
Leo Gao, Stella Biderman, Sid Black, Laurence Golding, Travis Hoppe, Charles Foster, Jason Phang, Horace He, Anish Thite, Noa Nabeshima, Shawn Presser, and Connor Leahy. 2021.
\newblock \href {https://arxiv.org/abs/2101.00027} {The {P}ile: An 800gb dataset of diverse text for language modeling}.
\newblock \emph{ArXiv preprint}, abs/2101.00027.

\bibitem[{Gardner et~al.(2024)Gardner, Perdomo, and Schmidt}]{t4}
Joshua~P Gardner, Juan~Carlos Perdomo, and Ludwig Schmidt. 2024.
\newblock \href {https://openreview.net/forum?id=WH5blx5tZ1} {Large scale transfer learning for tabular data via language modeling}.
\newblock In \emph{The Thirty-eighth Annual Conference on Neural Information Processing Systems}.

\bibitem[{Gururangan et~al.(2023)Gururangan, Wortsman, Gadre, Dave, Kilian, Shi, Mercat, Smyrnis, Ilharco, Jordan, Heckel, Dimakis, Farhadi, Shankar, and Schmidt}]{open_lm}
Suchin Gururangan, Mitchell Wortsman, Samir~Yitzhak Gadre, Achal Dave, Maciej Kilian, Weijia Shi, Jean Mercat, Georgios Smyrnis, Gabriel Ilharco, Matt Jordan, Reinhard Heckel, Alex Dimakis, Ali Farhadi, Vaishaal Shankar, and Ludwig Schmidt. 2023.
\newblock {OpenLM}: a minimal but performative language modeling (lm) repository.
\newblock \url{https://github.com/mlfoundations/open_lm}.

\bibitem[{{H}amborg et~al.(2017){H}amborg, {M}euschke, {B}reitinger, and {G}ipp}]{Hamborg2017}
{F}elix {H}amborg, {N}orman {M}euschke, {C}orinna {B}reitinger, and {B}ela {G}ipp. 2017.
\newblock {news-please}: {A} {G}eneric {N}ews {C}rawler and {E}xtractor.
\newblock In \emph{{P}roceedings of the 15th {I}nternational {S}ymposium of {I}nformation {S}cience}.

\bibitem[{Han et~al.(2025)Han, Jian, Hu, Liu, Wang, Fan, Ai, Huang, He, Yang, and You}]{infimmwebmath}
Xiaotian Han, Yiren Jian, Xuefeng Hu, Haogeng Liu, Yiqi Wang, Qihang Fan, Yuang Ai, Huaibo Huang, Ran He, Zhenheng Yang, and Quanzeng You. 2025.
\newblock Infimm-webmath-40b: Advancing multimodal pre-training for enhanced mathematical reasoning.

\bibitem[{Hendrycks et~al.(2021)Hendrycks, Burns, Basart, Zou, Mazeika, Song, and Steinhardt}]{hendrycks2020measuring}
Dan Hendrycks, Collin Burns, Steven Basart, Andy Zou, Mantas Mazeika, Dawn Song, and Jacob Steinhardt. 2021.
\newblock \href {https://openreview.net/forum?id=d7KBjmI3GmQ} {Measuring massive multitask language understanding}.
\newblock In \emph{9th International Conference on Learning Representations, {ICLR} 2021, Virtual Event, Austria, May 3-7, 2021}. OpenReview.net.

\bibitem[{Hoffmann et~al.(2022)Hoffmann, Borgeaud, Mensch, Buchatskaya, Cai, Rutherford, Casas, Hendricks, Welbl, Clark et~al.}]{chinchilla}
Jordan Hoffmann, Sebastian Borgeaud, Arthur Mensch, Elena Buchatskaya, Trevor Cai, Eliza Rutherford, Diego de~Las Casas, Lisa~Anne Hendricks, Johannes Welbl, Aidan Clark, and 1 others. 2022.
\newblock Training compute-optimal large language models.
\newblock In \emph{Advances in Neural Information Processing Systems (NeurIPS)}.
\newblock \url{https://arxiv.org/abs/2203.15556}.

\bibitem[{Kohlsch\"{u}tter et~al.(2010)Kohlsch\"{u}tter, Fankhauser, and Nejdl}]{boilerpipe}
Christian Kohlsch\"{u}tter, Peter Fankhauser, and Wolfgang Nejdl. 2010.
\newblock \href {https://doi.org/10.1145/1718487.1718542} {Boilerplate detection using shallow text features}.
\newblock In \emph{Proceedings of the Third ACM International Conference on Web Search and Data Mining}, WSDM '10, page 441–450, New York, NY, USA. Association for Computing Machinery.

\bibitem[{Lehmberg et~al.(2016)Lehmberg, Ritze, Meusel, and Bizer}]{wdccommonstables}
Oliver Lehmberg, Dominique Ritze, Robert Meusel, and Christian Bizer. 2016.
\newblock \href {https://doi.org/10.1145/2872518.2889386} {A large public corpus of web tables containing time and context metadata}.
\newblock In \emph{Proceedings of the 25th International Conference Companion on World Wide Web}, WWW '16 Companion, page 75–76, Republic and Canton of Geneva, CHE. International World Wide Web Conferences Steering Committee.

\bibitem[{Leonhardt et~al.(2020)Leonhardt, Anand, and Khosla}]{boilernet}
Jurek Leonhardt, Avishek Anand, and Megha Khosla. 2020.
\newblock \href {https://doi.org/10.1145/3366424.3383547} {Boilerplate removal using a neural sequence labeling model}.
\newblock In \emph{Companion Proceedings of the Web Conference 2020}, WWW '20, page 226–229, New York, NY, USA. Association for Computing Machinery.

\bibitem[{Li et~al.(2024)Li, Fang, Smyrnis, Ivgi, Jordan, Gadre, Bansal, Guha, Keh, Arora, Garg, Xin, Muennighoff, Heckel, Mercat, Chen, Gururangan, Wortsman, Albalak, Bitton, Nezhurina, Abbas, Hsieh, Ghosh, Gardner, Kilian, Zhang, Shao, Pratt, Sanyal, Ilharco, Daras, Marathe, Gokaslan, Zhang, Chandu, Nguyen, Vasiljevic, Kakade, Song, Sanghavi, Faghri, Oh, Zettlemoyer, Lo, El-Nouby, Pouransari, Toshev, Wang, Groeneveld, Soldaini, Koh, Jitsev, Kollar, Dimakis, Carmon, Dave, Schmidt, and Shankar}]{dclm}
Jeffrey Li, Alex Fang, Georgios Smyrnis, Maor Ivgi, Matt Jordan, Samir Gadre, Hritik Bansal, Etash Guha, Sedrick Keh, Kushal Arora, Saurabh Garg, Rui Xin, Niklas Muennighoff, Reinhard Heckel, Jean Mercat, Mayee Chen, Suchin Gururangan, Mitchell Wortsman, Alon Albalak, and 40 others. 2024.
\newblock \href {https://proceedings.neurips.cc/paper_files/paper/2024/file/19e4ea30dded58259665db375885e412-Paper-Datasets_and_Benchmarks_Track.pdf} {Datacomp-lm: In search of the next generation of training sets for language models}.
\newblock In \emph{Advances in Neural Information Processing Systems}, volume~37, pages 14200--14282. Curran Associates, Inc.

\bibitem[{Lozhkov et~al.(2024{\natexlab{a}})Lozhkov, Ben~Allal, von Werra, and Wolf}]{lozhkov2024fineweb-edu}
Anton Lozhkov, Loubna Ben~Allal, Leandro von Werra, and Thomas Wolf. 2024{\natexlab{a}}.
\newblock \href {https://huggingface.co/datasets/HuggingFaceFW/fineweb-edu} {Fineweb-edu}.

\bibitem[{Lozhkov et~al.(2024{\natexlab{b}})Lozhkov, Li, Allal, Cassano, Lamy-Poirier, Tazi, Tang, Pykhtar, Liu, Wei, Liu, Tian, Kocetkov, Zucker, Belkada, Wang, Liu, Abulkhanov, Paul, Li, Li, Risdal, Li, Zhu, Zhuo, Zheltonozhskii, Dade, Yu, Krauß, Jain, Su, He, Dey, Abati, Chai, Muennighoff, Tang, Oblokulov, Akiki, Marone, Mou, Mishra, Gu, Hui, Dao, Zebaze, Dehaene, Patry, Xu, McAuley, Hu, Scholak, Paquet, Robinson, Anderson, Chapados, Patwary, Tajbakhsh, Jernite, Ferrandis, Zhang, Hughes, Wolf, Guha, von Werra, and de~Vries}]{lozhkov2024starcoder}
Anton Lozhkov, Raymond Li, Loubna~Ben Allal, Federico Cassano, Joel Lamy-Poirier, Nouamane Tazi, Ao~Tang, Dmytro Pykhtar, Jiawei Liu, Yuxiang Wei, Tianyang Liu, Max Tian, Denis Kocetkov, Arthur Zucker, Younes Belkada, Zijian Wang, Qian Liu, Dmitry Abulkhanov, Indraneil Paul, and 47 others. 2024{\natexlab{b}}.
\newblock \href {https://arxiv.org/abs/2402.19173} {Starcoder 2 and the stack v2: The next generation}.
\newblock \emph{ArXiv preprint}, abs/2402.19173.

\bibitem[{Mahabadi et~al.(2025)Mahabadi, Satheesh, Prabhumoye, Patwary, Shoeybi, and Catanzaro}]{nemotronccmath}
Rabeeh~Karimi Mahabadi, Sanjeev Satheesh, Shrimai Prabhumoye, Mostofa Patwary, Mohammad Shoeybi, and Bryan Catanzaro. 2025.
\newblock Nemotron-cc-math: A 133 billion-token-scale high quality math pretraining dataset.

\bibitem[{{Meta AI}(2024)}]{llama3}
{Meta AI}. 2024.
\newblock Introducing meta llama 3: The most capable openly available llm to date.
\newblock \url{https://ai.meta.com/blog/meta-llama-3/}.

\bibitem[{Paster et~al.(2023)Paster, Santos, Azerbayev, and Ba}]{paster2023openwebmath}
Keiran Paster, Marco~Dos Santos, Zhangir Azerbayev, and Jimmy Ba. 2023.
\newblock \href {https://arxiv.org/abs/2310.06786} {Openwebmath: An open dataset of high-quality mathematical web text}.
\newblock \emph{ArXiv preprint}, abs/2310.06786.

\bibitem[{Pasupat and Liang(2016)}]{pasupat-liang-2016-inferring}
Panupong Pasupat and Percy Liang. 2016.
\newblock \href {https://doi.org/10.18653/v1/P16-1003} {Inferring logical forms from denotations}.
\newblock In \emph{Proceedings of the 54th Annual Meeting of the Association for Computational Linguistics (Volume 1: Long Papers)}, pages 23--32, Berlin, Germany. Association for Computational Linguistics.

\bibitem[{Penedo et~al.(2023)Penedo, Malartic, Hesslow, Cojocaru, Cappelli, Alobeidli, Pannier, Almazrouei, and Launay}]{refinedweb}
Guilherme Penedo, Quentin Malartic, Daniel Hesslow, Ruxandra Cojocaru, Alessandro Cappelli, Hamza Alobeidli, Baptiste Pannier, Ebtesam Almazrouei, and Julien Launay. 2023.
\newblock \href {https://arxiv.org/abs/2306.01116} {The {R}efined{W}eb dataset for {F}alcon {LLM}: outperforming curated corpora with web data, and web data only}.
\newblock \emph{ArXiv preprint}, abs/2306.01116.

\bibitem[{Pomik{\'a}lek(2011)}]{justext1}
Jan Pomik{\'a}lek. 2011.
\newblock \emph{Removing boilerplate and duplicate content from web corpora}.
\newblock Ph.D. thesis, Masaryk university, Faculty of informatics, Brno, Czech Republic.

\bibitem[{Pouransari et~al.(2024)Pouransari, Li, Chang, Vasu, Koc, Shankar, and Tuzel}]{pouransari2024dataset}
Hadi Pouransari, Chun-Liang Li, Jen-Hao~Rick Chang, Pavan Kumar~Anasosalu Vasu, Cem Koc, Vaishaal Shankar, and Oncel Tuzel. 2024.
\newblock \href {arXiv preprint arXiv:2405.13226} {Dataset decomposition: Faster llm training with variable sequence length curriculum}.
\newblock \emph{arXiv preprint arXiv:2405.13226}.

\bibitem[{Raffel et~al.(2020)Raffel, Shazeer, Roberts, Lee, Narang, Matena, Zhou, Li, and Liu}]{c4}
Colin Raffel, Noam Shazeer, Adam Roberts, Katherine Lee, Sharan Narang, Michael Matena, Yanqi Zhou, Wei Li, and Peter~J. Liu. 2020.
\newblock \href {http://jmlr.org/papers/v21/20-074.html} {Exploring the limits of transfer learning with a unified text-to-text transformer}.
\newblock \emph{J. Mach. Learn. Res.}, 21:140:1--140:67.

\bibitem[{Soldaini et~al.(2024)Soldaini, Kinney, Bhagia, Schwenk, Atkinson, Authur, Bogin, Chandu, Dumas, Elazar et~al.}]{soldaini2024dolma}
Luca Soldaini, Rodney Kinney, Akshita Bhagia, Dustin Schwenk, David Atkinson, Russell Authur, Ben Bogin, Khyathi Chandu, Jennifer Dumas, Yanai Elazar, and 1 others. 2024.
\newblock \href {https://arxiv.org/abs/2402.00159} {Dolma: An open corpus of three trillion tokens for language model pretraining research}.
\newblock \emph{ArXiv preprint}, abs/2402.00159.

\bibitem[{Su et~al.(2024)Su, Kong, Lin, Jennings, Norick, Kliegl, Patwary, Shoeybi, and Catanzaro}]{su2024nemotroncc}
Dan Su, Kezhi Kong, Ying Lin, Joseph Jennings, Brandon Norick, Markus Kliegl, Mostofa Patwary, Mohammad Shoeybi, and Bryan Catanzaro. 2024.
\newblock Nemotron-cc: Transforming common crawl into a refined long-horizon pretraining dataset.
\newblock \emph{arXiv preprint arXiv:2412.02595}.

\bibitem[{{Together Computer}(2023)}]{rpj}
{Together Computer}. 2023.
\newblock Redpajama: an open dataset for training large language models.
\newblock \url{https://github.com/togethercomputer/RedPajama-Data}.

\bibitem[{van Breugel and van~der Schaar(2024)}]{vanbreugel2024}
Boris van Breugel and Mihaela van~der Schaar. 2024.
\newblock Why tabular foundation models should be a research priority.
\newblock \emph{arXiv preprint arXiv:2405.01147}.

\bibitem[{Vogels et~al.(2018)Vogels, Ganea, and Eickhof}]{web2text}
Thijs Vogels, Octavian-Eugen Ganea, and Carsten Eickhof. 2018.
\newblock Web2text: Deep structured boilerplate removal.
\newblock In \emph{Advances in Information Retrieval, ECIR 2018}.

\bibitem[{Wei et~al.(2024)Wei, Cassano, Liu, Ding, Jain, de~Vries, von Werra, Guha, and Zhang}]{starcoder2instruct}
Yuxiang Wei, Federico Cassano, Jiawei Liu, Yifeng Ding, Naman Jain, Harm de~Vries, Leandro von Werra, Arjun Guha, and Lingming Zhang. 2024.
\newblock Starcoder2-instruct: Fully transparent and permissive self-alignment for code generation.
\newblock \url{https://huggingface.co/blog/sc2-instruct}.

\bibitem[{Wortsman et~al.(2022)Wortsman, Ilharco, Kim, Li, Kornblith, Roelofs, Lopes, Hajishirzi, Farhadi, Namkoong, and Schmidt}]{wiseft}
Mitchell Wortsman, Gabriel Ilharco, Jong~Wook Kim, Mike Li, Simon Kornblith, Rebecca Roelofs, Raphael~Gontijo Lopes, Hannaneh Hajishirzi, Ali Farhadi, Hongseok Namkoong, and Ludwig Schmidt. 2022.
\newblock Robust fine-tuning of zero-shot models.
\newblock In \emph{Proceedings of the IEEE/CVF Conference on Computer Vision and Pattern Recognition (CVPR)}, pages 7959--7971.

\bibitem[{Xu et~al.(2024)Xu, Liu, Yan, Liu, Xiong, and Yu}]{xu2024cleaner}
Zhipeng Xu, Zhenghao Liu, Yukun Yan, Zhiyuan Liu, Chenyan Xiong, and Ge~Yu. 2024.
\newblock Cleaner pretraining corpus curation with neural web scraping.
\newblock In \emph{Proceedings of the 62nd Annual Meeting of the Association for Computational Linguistics}.

\bibitem[{Yin et~al.(2020)Yin, Neubig, tau Yih, and Riedel}]{tabert}
Pengcheng Yin, Graham Neubig, Wen tau Yih, and Sebastian Riedel. 2020.
\newblock Ta{BERT}: Pretraining for joint understanding of textual and tabular data.
\newblock In \emph{Annual Conference of the Association for Computational Linguistics (ACL)}.

\bibitem[{Zamazal(2024)}]{justext2}
Kryštof Zamazal. 2024.
\newblock Evaluation of web page cleaning tool justext.

\bibitem[{Zhang and Balog(2025)}]{zhang2020webtableextractionretrieval}
Shuo Zhang and Krisztian Balog. 2025.
\newblock Web table extraction, retrieval and augmentation: A survey.
\newblock \emph{arXiv preprint arXiv:2002.00207}.

\end{thebibliography}

\newpage
\appendix

\clearpage

\section{Extended related work}\label{app:extended_related_work}

\paragraph{Targeted extractors for specific domains.} When extracting specific types of web pages, other works have sometimes developed custom approaches. For news articles, \texttt{newspaper3k}\footnote{\url{https://github.com/codelucas/newspaper}} and \texttt{news-please}~\citep{Hamborg2017} target cleaner extractions with better date and author parsing. For cooking recipes, \texttt{recipe-scrapers}\footnote{\url{https://github.com/hhursev/recipe-scrapers}} offers targeted parsing of ingredients, instructions, and cooking times.  More recently, in the context of pretraining, custom extractors have been heavily emphasized as an important step for building mathematics datasets. As more general approaches were shown to inaccurately/inconsistently handle LaTeX equations, both OpenWebMath~\citep{paster2023openwebmath} and InfiMM-WebMath~\citep{infimmwebmath} develop their own modifications to \resiliparse{}, the former of which was also later used by FineMath~\citep{smollm2}. Nemotron-CC-Math~\citep{nemotronccmath} offers state-of-the-art performance by re-extracting pages from the aforementioned datasets by making use of \texttt{lynx}, a text-based browser that more robustly handles math and code. They also clean these extractions with a fine-tuned LLM. 

\paragraph{Datasets for web-crawled tables} Compared to simply collecting existing tabular datasets, a promising complementary approach for improving tabular understanding is to curate large interleaved table/text datasets from the general web \cite{zhang2020webtableextractionretrieval}. TabLib~\citep{tablib} collects 627M (minimally filtered) tables with their contextual metadata (e.g., surrounding text) by crawling GitHub and one dump of Common Crawl (\texttt{CC-MAIN-2023-23}). Of their tables, 219M are HTML tables from Common Crawl. Tablib was then further filtered to create T4~\citep{t4} via a set of table/row/column-level heuristics (e.g., based on counts, prevalence of NANs, and data homogeniety). The goal of this dataset was to improve downstream model performance for tabular prediction tasks (i.e., selecting and predicting a chosen target feature from the other feature columns). Another set of earlier but closely related efforts are the web tables collected by Web Data Commons (WDC)~\citep{wdccommonstables, dresdentables}. These datasets are created from a pair of Common Crawl dumps (from 2012 and 2015) and involve a two step process where tables are filtered based on (1) hand-engineered features (e.g., sparseness, average attribute size, number of links); (2) classifying each table as either Relational, Entity, Matrix, or Layout via learned classifiers. The relational subset was used by \citet{tabert} to train TaBERT (along with an additional dataset based on just Wikipedia tables). In comparison, our approach for CC-Tables uses a smaller set of heuristic rules (i.e., based on header presence, row/columns counts, and row size consistency) and trains a weakly supervised \fasttext{} classifier directly on HTML content rather than a SVM or decision tree on hand-engineered features. 

\paragraph{Datasets for interleaved code and natural text.} For open code pretraining datasets, The Stack v2 and its filtered subsets from StarCoder~\citep{lozhkov2024starcoder} are commonly used directly to train models or as starting points for further curation. Mostly created from GitHub, it primarily consists of source code but also contains some with interleaved code/text (e.g., Issues/Commits/PRs, Jupyter notebooks, and code documentation). Meanwhile, RedStone-Code~\citep{redstone} directly sources interleaved documents by filtering pages from Common Crawl. They focus on \texttt{<code>} elements, classify blocks as code (or not) based upon regexes for common language keywords/patterns, and extract such pages using the same pipeline as Common Crawl WET files. In contrast, we train a \fasttext{} classifier instead of using regex rules and we do not rely on the presence of \texttt{<code>}, though we do use it as signal for sourcing positive training examples for our filtering model. Lastly, Nemotron-CC-Math~\citep{nemotronccmath}, as mentioned earlier, re-extracts content from popular math datasets (which incidentally contains significant amounts of code), finding that their re-extractions can improve performance on code evaluations.

\clearpage

\section{Assets and licenses}

We used the following assets all for research purposes. \\

\noindent \textbf{Datasets} \begin{itemize}
    \item We make heavy use of the DCLM codebase and datasets (\rawpool, \dclmrw, \hqpool), which are available by CC-BY-4 license.
    \item We also directly source raw data from Common Crawl WARCs, which are under their own ToU\footnote{\url{https://commoncrawl.org/terms-of-use}}.
\end{itemize} 

\noindent \textbf{Extractor libraries} \begin{itemize}
    \item \resiliparse{} \texttt{(v0.14.5)} is made available via Apache 2.0 License. We use this version to match up with DCLM.
    \item \trafilatura{} \texttt{(v2.0.0)} is made available via Apache 2.0 License
    \item \justext{} \texttt{(v3.0.2)} is made available via The BSD 2-Clause License
\end{itemize}

\noindent \textbf{Evaluation harnesses} \begin{itemize}
    \item \texttt{llm-foundry}\footnote{\url{https://github.com/mosaicml/llm-foundry}} is used for DCLM's evaluations and is available under Apache 2.0 License.
    \item \texttt{unitxt}\footnote{\url{https://github.com/IBM/unitxt}}~\citep{bandel-etal-2024-unitxt} implements ToRR~\citep{torr} and is available via Apache 2.0 License
    \item \texttt{bigcode-evaluation-harness}\footnote{\url{https://github.com/bigcode-project/bigcode-evaluation-harness}}~\citep{bigcode-evaluation-harness} is avaialble under Apache 2.0 License
\end{itemize}

\clearpage

\onecolumn
\section{Additional results for union over extractors} \label{app:union_extended}

\begin{table}[h!]
    \centering
    \small
    \setlength{\tabcolsep}{4pt}
     \caption{\textbf{Run-to-run variation at \strack{} scale.} We report mean \lowvar{} performance and standard deviation across three runs for \resiliparse{} and \trafilatura{}. Note that results in \Cref{tab:extractor_union} were all single-runs since we did not try multiple seeds for all methods.}
    \begin{tabular}{c|ccc|c}
    \toprule
    Extractor & Run 1 & Run 2 & Run 3 & Mean {\scriptsize\color{gray}(Std)} \\
    \midrule
    \resiliparse{} & 28.5 & 27.5 & 28.8 & 28.3 {\scriptsize\color{gray}(0.7)} \\
    \trafilatura{} & 29.6 & 27.7 & 28.5 & 28.6 {\scriptsize\color{gray}(1.0)} \\
    \bottomrule
    \end{tabular}
    \label{tab:run_variation}
\end{table}

\begin{table*}[h!]
    \centering
    \small
    \caption{\textbf{Using multiple extractors improves performance in settings which are data-constrained.}
    We train \texttt{1B-5x} models on progressively smaller subsamples of data curated from the \strack{} raw pool. More heavy subsampling leads to more repetition and thus worse performance, but this can be mitigated by having a curation pipeline that yields more tokens (e.g., our Union datasets). A subset of these results are plotted in \Cref{fig:union_data_constrained}.}
    \begin{tabular}{c|lcccc}
    \toprule
    Initial Pool & Extractor & \fasttext{} Thresh. & Repeats & \lowvar \\
    \bottomrule
          \multirow{5}{*}{\texttt{1B-0.25x}} & \resiliparse{}  & 0.11 &  15.0x & 30.8  \\
      & Union (Random)  & (0.11, 0.11, 0.11) &  10.5x & 32.3  \\
      & Union (Manual)  & (0.11, 0.11, 0.11) &  10.2x & 32.4 \\
      & Union (Random)  & (0.11, 0.15, 0.15) &  9.0x & \textbf{33.5}  \\
      & Union (Manual)  & (0.11, 0.15, 0.15) &  8.7x & \textbf{33.5} \\
      \hline
      \multirow{5}{*}{\texttt{1B-0.33x}} & \resiliparse{}  & 0.11 & 11.2x & 33.1 \\
      & Union (Random)  & (0.11, 0.11, 0.11) & 7.9x & 33.3 \\
      & Union (Manual)  & (0.11, 0.11, 0.11) & 7.7x & 33.3 \\
      & Union (Random)  & (0.11, 0.15, 0.15) & 6.6x & 33.4 \\
      & Union (Manual)  & (0.11, 0.15, 0.15) & 6.3x & \textbf{33.8}\\
      \hline
      \multirow{5}{*}{\texttt{1B-0.5x}} & \resiliparse{}  & 0.11 & 7.2x & 32.7 \\
      & Union (Random)  & (0.11, 0.11, 0.11) & 5.3x & 34.0 \\
      & Union (Manual)  & (0.11, 0.11, 0.11) & 5.1x & 34.3\\
      & Union (Random)  & (0.11, 0.15, 0.15) & 4.4x & \textbf{34.6} \\
      & Union (Manual)  & (0.11, 0.15, 0.15) & 4.2x & 34.3 \\
      \hline
      \multirow{5}{*}{\texttt{1B-1x}} & \resiliparse{}  & 0.11 & 3.6x & 33.9 \\
      & Union (Random)  & (0.11, 0.11, 0.11) & 2.6x & 34.0\\
      & Union (Manual)  & (0.11, 0.11, 0.11) & 2.5x & 33.7 \\
      & Union (Random)  & (0.11, 0.15, 0.15) & 2.2x & \textbf{34.4} \\
      & Union (Manual)  & (0.11, 0.15, 0.15) & 2.1x & 33.8 \\ \bottomrule
\end{tabular}  
    \label{tab:union_data_constrained}
\end{table*}

\clearpage
\section{Table understanding experiments}\label{app:tables}

\subsection{Example documents}\label{app:tab_examples}

\vspace{6mm}

\noindent\centering\small\textbf{Example 1:} \url{http://mutualfunds.com/funds/prblx-parnassus-core-equity-investor/}
\begin{figure}[h!]
    \centering
    \includegraphics[width=1.0\linewidth]{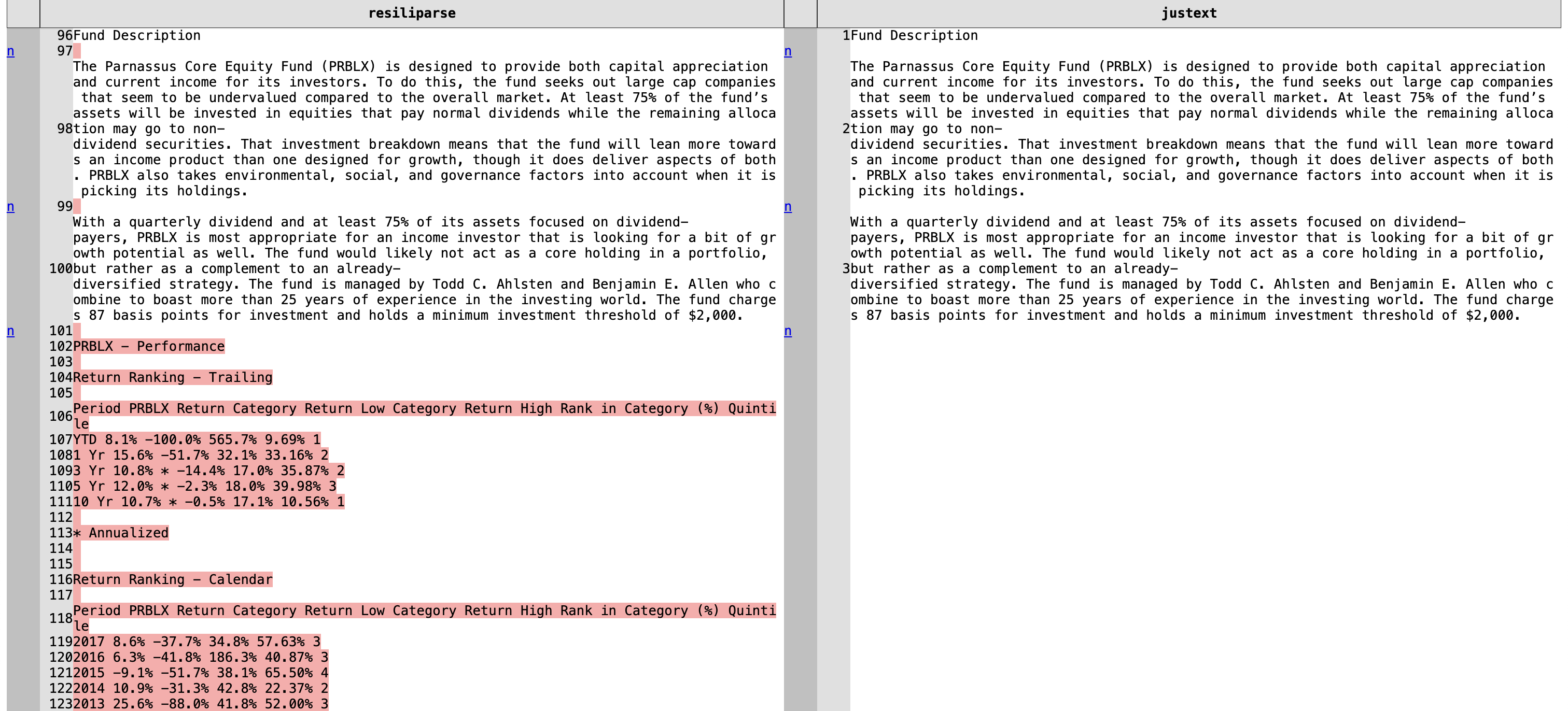}

    \vspace{6mm}

    \includegraphics[width=1.0\linewidth]{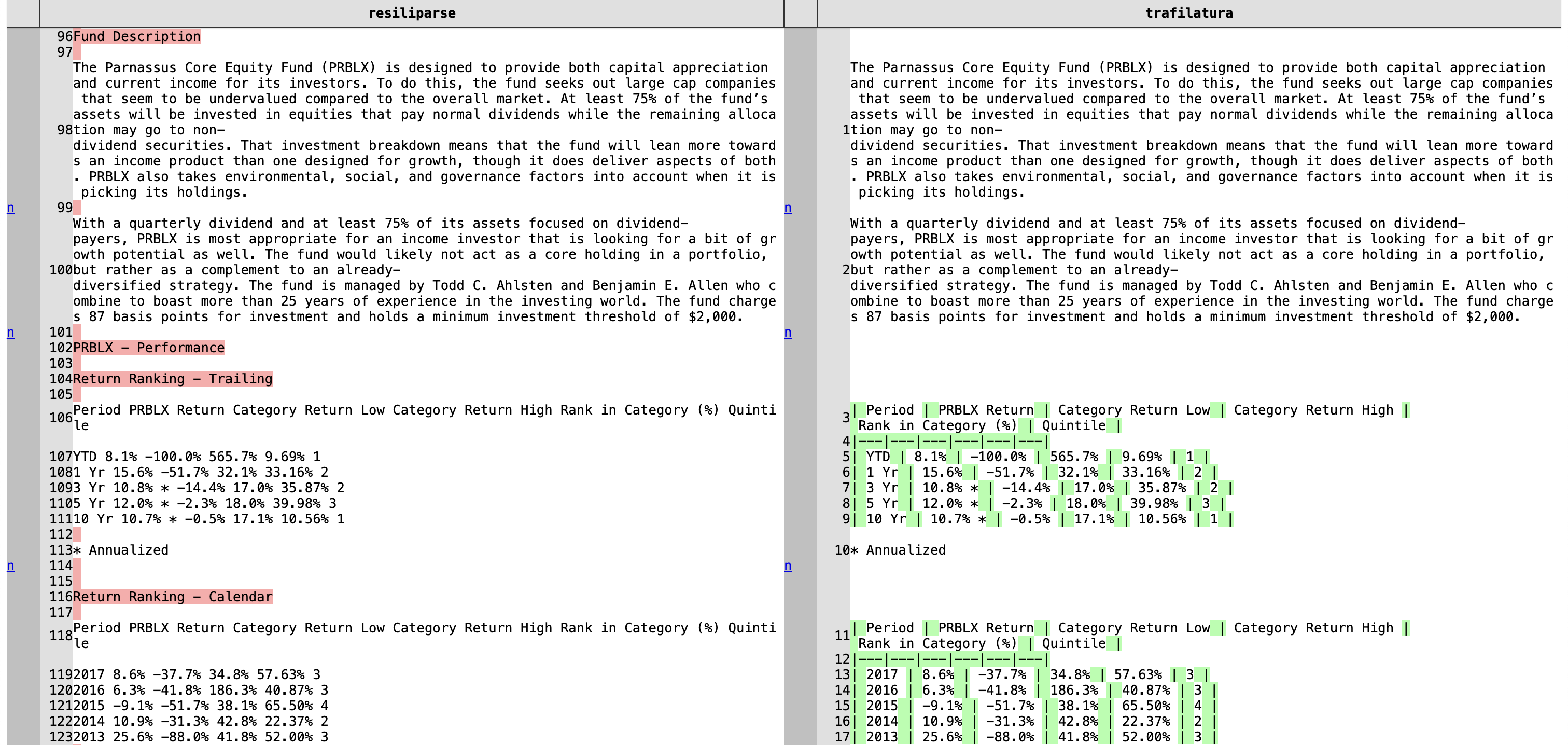}
    \caption{\textbf{Extraction comparison for mutual fund data.} We use \texttt{difflib} to visualize pairwise comparisons between \resiliparse{} (left) and \justext{} (top) or \trafilatura{} (bottom). For both tables in this page, \justext{} removes them while \trafilatura{} applies markdown formatting.}
    \label{fig:table_example_1}
\end{figure}

\clearpage
\noindent\centering\small\textbf{Example 2:} \url{http://bmcinfectdis.biomedcentral.com/articles/10.1186/1471-2334-12-102}
\begin{figure}[h!]
    \centering
    \includegraphics[width=1.0\linewidth]{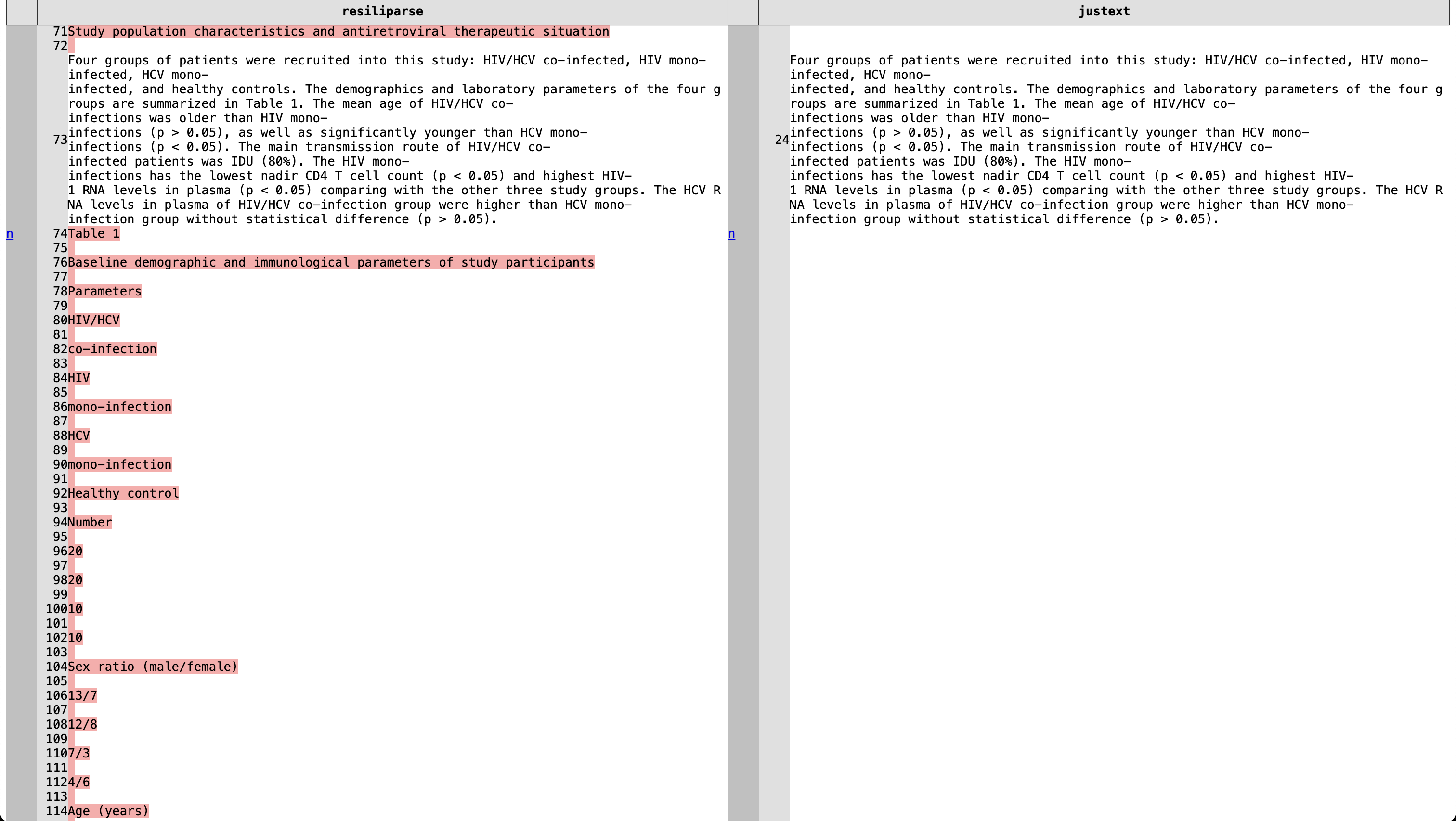}

    \vspace{6mm}

    \includegraphics[width=1.0\linewidth]{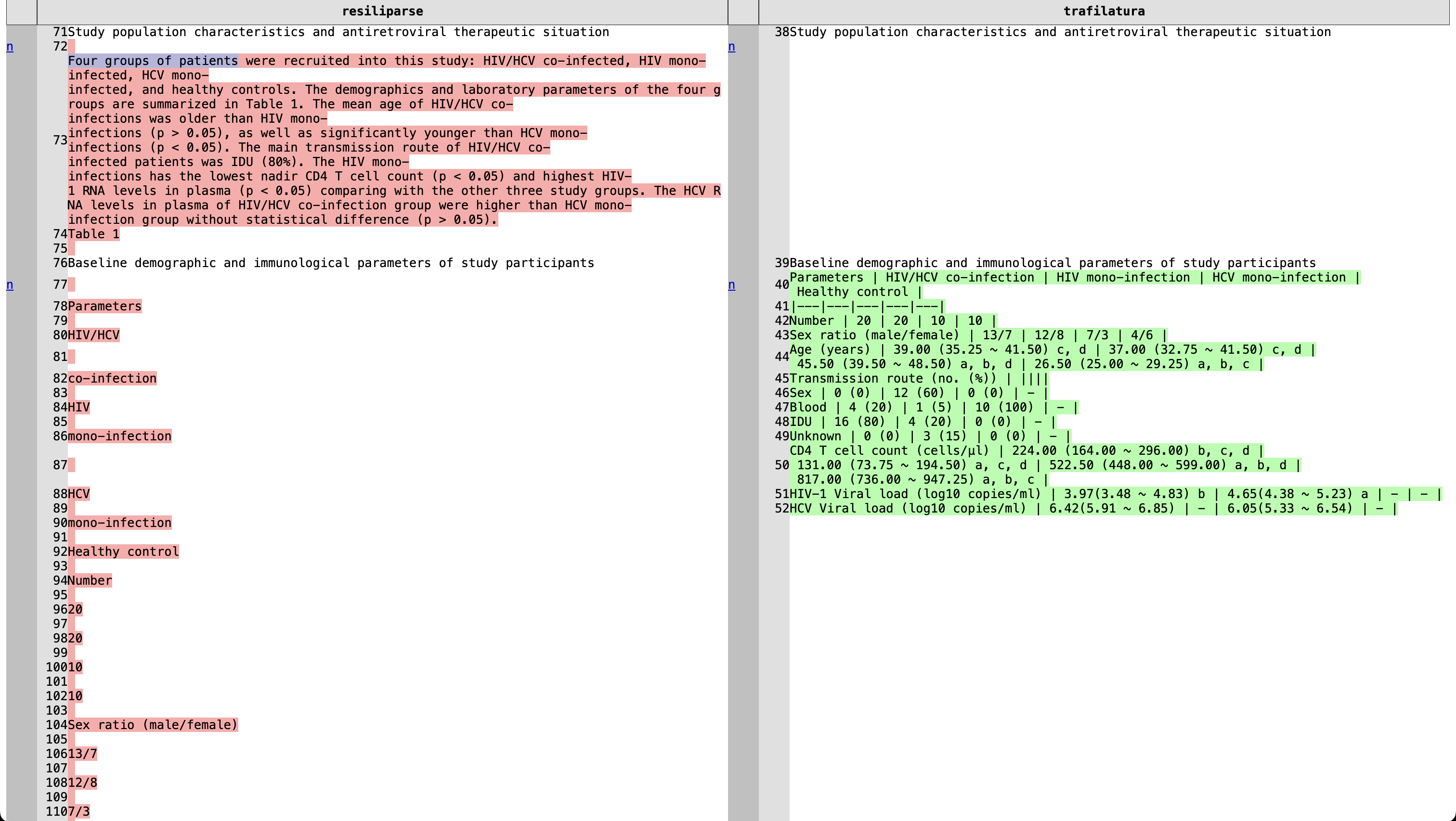}
    \caption{\textbf{Extraction comparison for an BioMed Central article.} We use \texttt{difflib} to visualize pairwise comparisons between \resiliparse{} (left) and \justext{} (top) or \trafilatura{} (bottom). Note that for the table shown, \justext{} removes it while \trafilatura{} applies markdown formatting. Here, \resiliparse{} ends up splitting entries across line breaks instead of single spaces.}
    \label{fig:table_example_2}
\end{figure}

\clearpage
\noindent\centering\small\textbf{Example 3}: \url{http://akron.prestosports.com/sports/msoc/2013-14/bios/caso%20clint%20yxrk}
\begin{figure}[h!]
    \centering
    \includegraphics[width=1.0\linewidth]{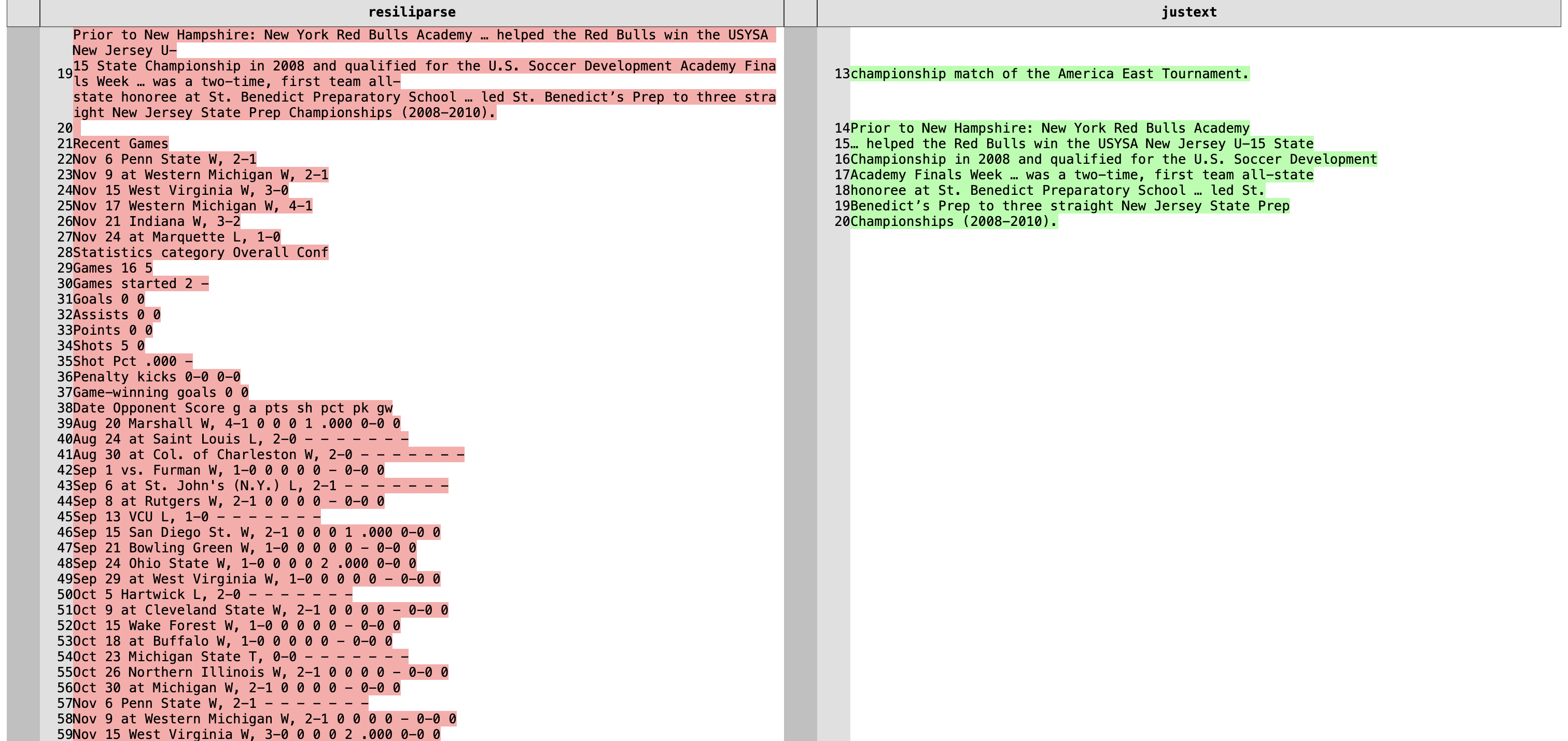}

    \vspace{6mm}

    \includegraphics[width=1.0\linewidth]{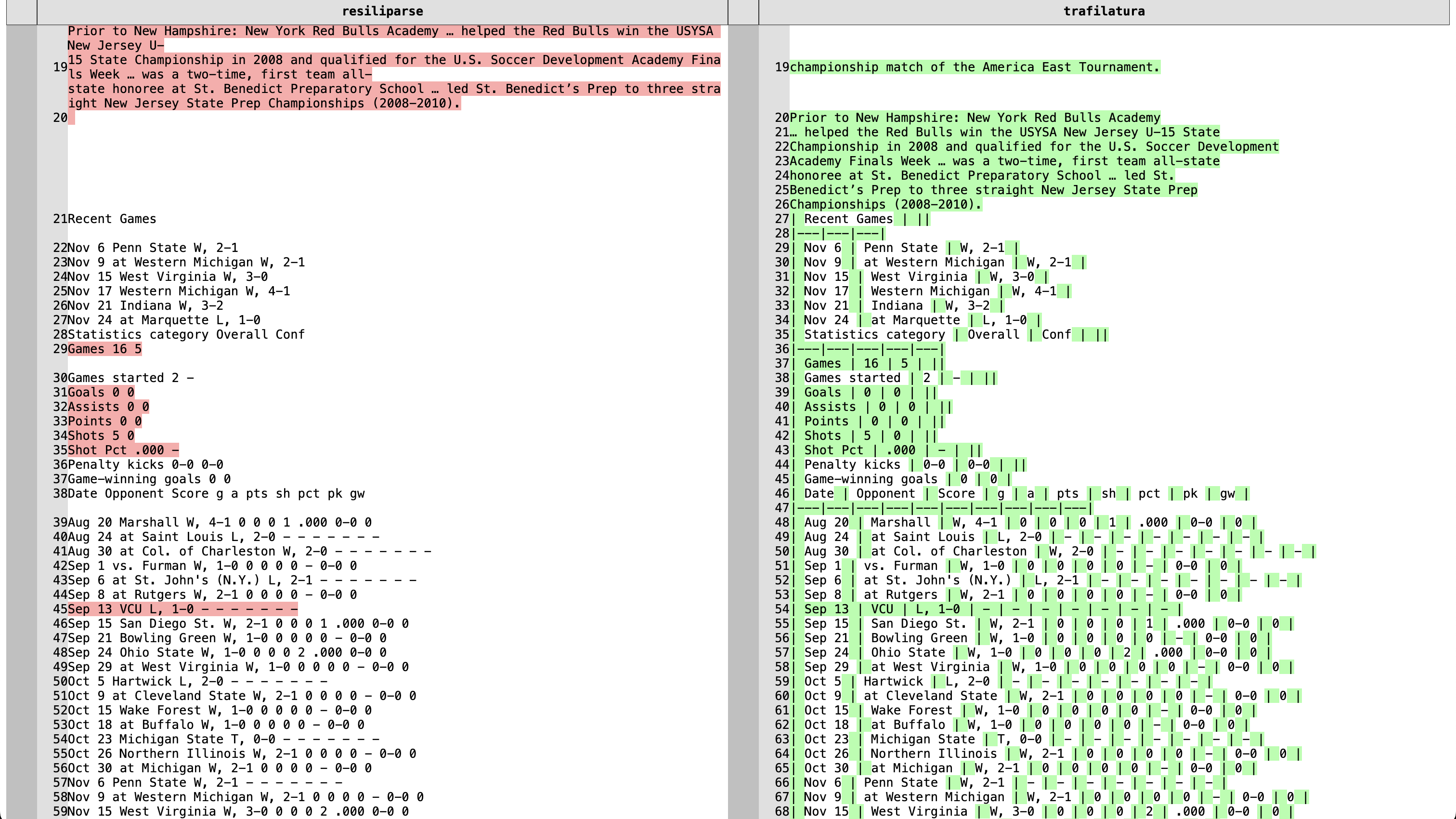}
    \caption{\textbf{Extraction comparison for soccer recruiting statistics.} We use \texttt{difflib} to visualize pairwise comparisons between \resiliparse{} (left) and \justext{} (top) or \trafilatura{} (bottom). Note that for both tables in this page, \justext{} removes them while \trafilatura{} applies markdown formatting.}
    \label{fig:table_example_3}
\end{figure}

\clearpage
\noindent {\centering\small\textbf{Example 4:}} \small{\url{https://bulbapedia.bulbagarden.net/wiki/Whiscash\_(Pok%C3%A9mon)}}

\begin{figure}[h!]
    \centering

    \includegraphics[width=1.0\linewidth]{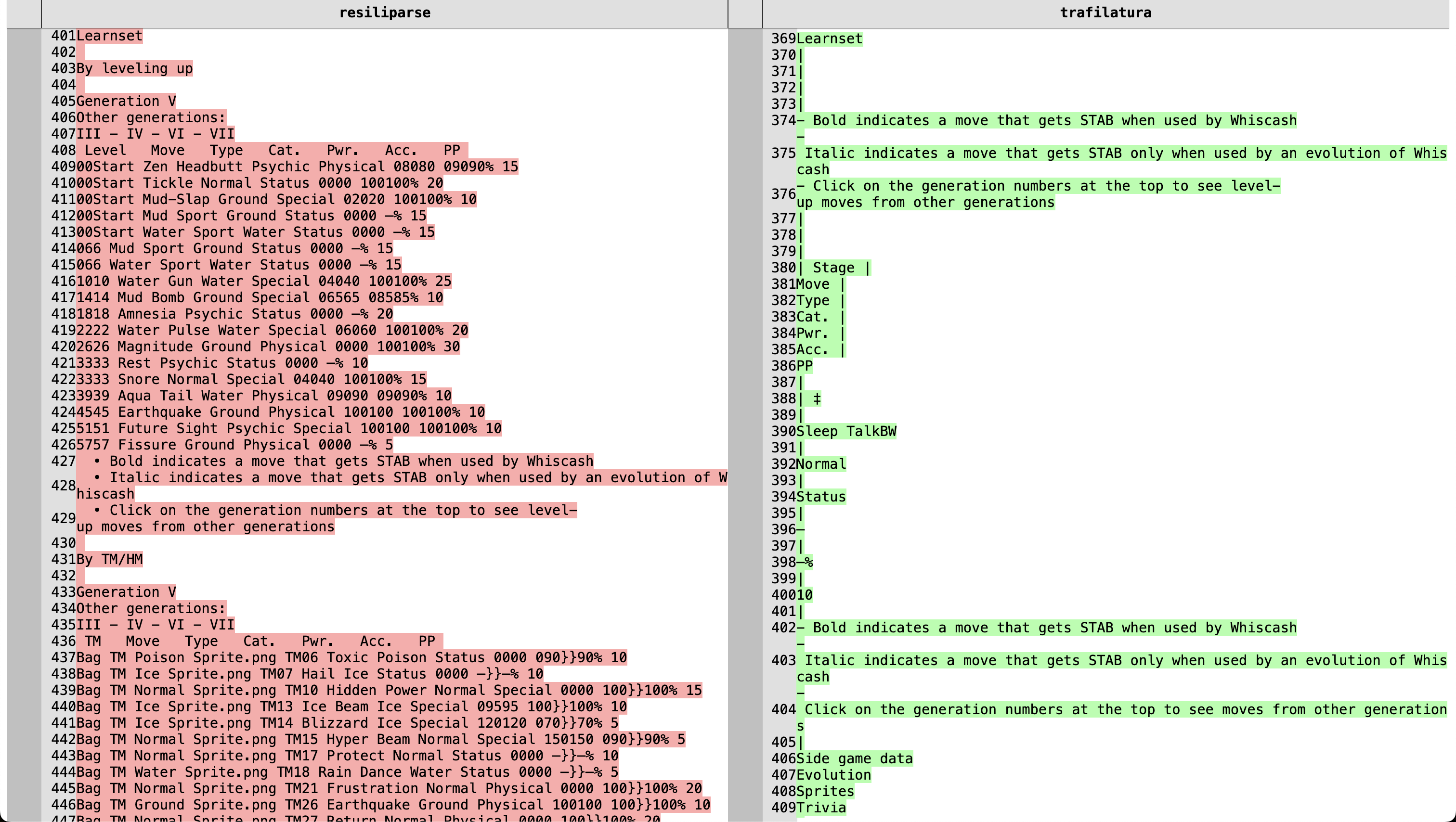}
    \caption{\textbf{Extraction comparison for a Pokémon learnset table on Bulbapedia.} We use \texttt{difflib} to visualize pairwise comparisons between \resiliparse{} and \trafilatura{}. Note that \justext{} removes the tables altogether so we do not show it. Meanwhile, \resiliparse{} fails to cleanly split some columns while \trafilatura{} fails to keep the actual contents.}
    \label{fig:table_example_4}
\end{figure}

\clearpage

\normalsize \justifying  \twocolumn

\subsection{Filtering for CC-Tables}\label{app:table_filtering} 

\normalsize We start from the raw data corresponding to the \xltrack{} version of \dclmrw. Here though, we prioritize filtering for pages with useful tables rather than for general quality, starting by considering only pages that contain \texttt{<table>} elements. Then we isolate these tables and filter them with a two-stage pipeline to try to keep genuine (relational) data tables. This is important as \texttt{<table>} is often used for other purposes such as to organize text in an aligned layout. We assume these ``non-genuine'' tables would not be useful for tabular understanding tasks. \newline

\noindent \textbf{Structure-based filtering.} In this first stage, we filter out any table that \begin{itemize}
    \item Does not contain headers (i.e., \texttt{<th>} elements)
    \item Has fewer than 10 rows and 3 columns. These are more strict thresholds than those used by~\citet{wdccommonstables}. 
    \item Has an inconsistent number of columns per row. 
\end{itemize}

\noindent \textbf{Content-based filtering.} After this first stage, we still found that many tables appeared that were unlikely to be useful (e.g. product lists, sizing charts, lists of forum threads). We also wished to prioritize tables that exist in knowledge-rich pages that interleave tables with natural text (e.g., Wikipedia articles). Based on these two goals, we train a \fasttext{} model to classify and further filter HTML tables.  To do so, we first inspected the tables that survived structure-based filtering and created a weakly labeled training set. Specifically, we source positive and negative examples by inspecting tables and observing patterns from their source page's URLs\footnote{Note that while we make use of URLs for labeling, we do not use them as input features to the model.}: \newline

\noindent \textit{Negative examples:} We focus on tables whose contents are either empty, esoteric, or unlikely to have supporting natural text   \begin{itemize}
    \item \textit{Product listings}: URLs containing one of the subwords: \texttt{``shop'', ``products'', ``cart'', ``items'', ``store'', ``promotion'', ``productdisplay''}
    \item \textit{Site metrics}: URLs that end with \texttt{``metrics''}
    \item \textit{Forum listings}: URLs containing either \texttt{``forum''} or \texttt{``forums''} as a subword
    \item \textit{User profile pages}: URLs containing one of the subwords: \texttt{``users'', ``members-''}
    \item \textit{Weather forecasts from Accuweather:} URLs from \texttt{``accuweather\.com''} 
    \item \textit{Patent Listings:} URLs from Google Patents as they contain little to no associated natural text
\end{itemize}

\noindent \textit{Positive examples:} We focus on genuine relational data tables that are likely to contain useful knowledge and be associated with natural text on their source page. \begin{itemize}
    \item \textit{Wikipedia}: URLs from English Wikipedia
    \item \textit{Government websites:} URLs from \texttt{``.gov''} domains
    \item \textit{Dataset releases:} URLs containing one of the subwords: \texttt{``statistics'',  ``database'', ``dataset''}
    \item \textit{Article and Documentation:} URLs containing the subwords \texttt{``article''} or substring \texttt{``docs.''} to encourage interleaved text and tables 
    \item Tables from T4: These are tables from TabLib~\citep{tablib} that have been filtered as higher-quality by \citet{t4}. While not all of these tables originate from Common Crawl (and were used by \citet{t4} as training data for tabular prediction tasks), we convert them to HTML serializations using \texttt{pandas}. We believe these tables are useful to include both for their contents but also the clean resulting HTML structure.  
\end{itemize}

We then train a \fasttext{} model on the HTML versions of these tables, allowing the model to use both contents and markup. While by default \fasttext{} uses space-delimiting for constructing n-grams, we found it to be better to first tokenize each HTML table using the \texttt{o200k\_base} encoder from \texttt{tiktoken}, allowing the model to instead train and make predictions on sequences of token indices.  We use default hyperparameters with the exception of increasing the number of n-grams to 4 (from the default of 1). When applying our \fasttext{} model, we consider any table to have a score above 0.75 to be useful. We then remove any page that does not contain a single table that meets this threshold. This left us with 62B tokens after \resiliparse{} extraction, 59B tokens after \trafilatura{}, and 21B tokens after \justext{}.

\subsection{Modifications to ToRR's WikiTQ  }\label{app:table_eval}

In the original ToRR~\citep{torr} implementation of WikiTQ, the chosen preprocessors convert both the generated predictions and answers to the string representation of a list of words. This ends up rewarding short generations even if they contain a completely incorrect answer. \newline

As an example, suppose \texttt{<prediction>} and \texttt{<answer>} are single-token words: 

\begin{itemize}
    \item They get converted to the strings \texttt{``\green{[‘<prediction>’]}''} and \texttt{``\green{[‘<answer>’]}''}
    
    \item The evaluation uses an F1 score that is applied over the tokenized sets of 5 tokens each: \begin{itemize}
        \item Prediction:  \texttt{\{ \green{[}, \green{‘}, \green{<prediction>}, \green{'}, \green{]}\}}  
        \item Answer: \texttt{\{ \green{[}, \green{`}, \green{<answer>}, \green{'}, \green{]} \}}
    \end{itemize}
    
    \item The brackets and single-quotes count as tokens and always appear in both the prediction and answer being compared. So even when \texttt{<prediction>} $\neq$ \texttt{<answer>}, we’d still get a score of 0.8.
\end{itemize}

To address this, we removed the list conversion (which is unnecessary since each true answer refers to a single entity) and added standard normalizers for leading/trailing whitespace and lowercasing. We now directly compute F1 over prediction and answer strings (tokenized with SpaCy).
\clearpage

\onecolumn
\subsection{Additional results}\label{app:tab_additional_results}

\begin{table*}[h!]
    \centering     
    \small

    \caption{\textbf{Conclusions about extractors hold regardless of mixing ratio or whether the \hqpool and CC-Tables extractors are consistent (\ltrack{} scale).} We compare the performance of \resiliparse{}, \trafilatura{}, and \justext{} on pages from our CC-Tables dataset when it is mixed with different ratios and different extractions of \hqpool (bottom section). Overall, we observe that there is a slight increase to WikiTQ performance for \resiliparse{} as the ratio of CC-Tables is larger but that conclusions between the extractors are the same across all settings.  }
    \begin{tabular}{ccccc}
    \toprule
    CC-Tables Extraction & \hqpool Extraction  & CC-Tables Mixing Ratio  & WikiTQ-Avg.  \\
    \midrule
    \resiliparse{} & \resiliparse{}   & 0.1 & \textbf{10.1} \\
    \trafilatura{} & \resiliparse{}  & 0.1 & 2.2 \\
    \justext{} & \resiliparse{}   & 0.1 & 1.9 \\  
    \midrule
    \resiliparse{} & \resiliparse{} &  0.2 & \textbf{11.9}  \\
    \trafilatura{} & \resiliparse{}  & 0.2 &  3.7\\
    \justext{} & \resiliparse{}   & 0.2 &  1.6 \\ \midrule
    \resiliparse{} & \resiliparse{} &  0.5 & \textbf{12.4}  \\
    \trafilatura{} & \resiliparse{}  & 0.5 & 3.4  \\
    \trafilatura{} & \trafilatura{}  & 0.5 &  2.7 \\
    \justext{} & \resiliparse{}   & 0.5 &  1.5  \\ 
    \justext{} & \justext{}   & 0.5 &  1.7  \\ 
    \bottomrule
    \end{tabular}  
    \label{tab:tables_appendix}
\end{table*}

\clearpage

\onecolumn
\section{Code experiments}
\subsection{Example documents}\label{app:code_examples}

\vspace{6mm}

\noindent\centering\small\textbf{Example 1:} \url{http://msdn.microsoft.com/en-us/library/cx9s2sy4(v=vs.100).aspx}

\begin{figure}[h!]
    \centering
    \includegraphics[width=1.0\linewidth]{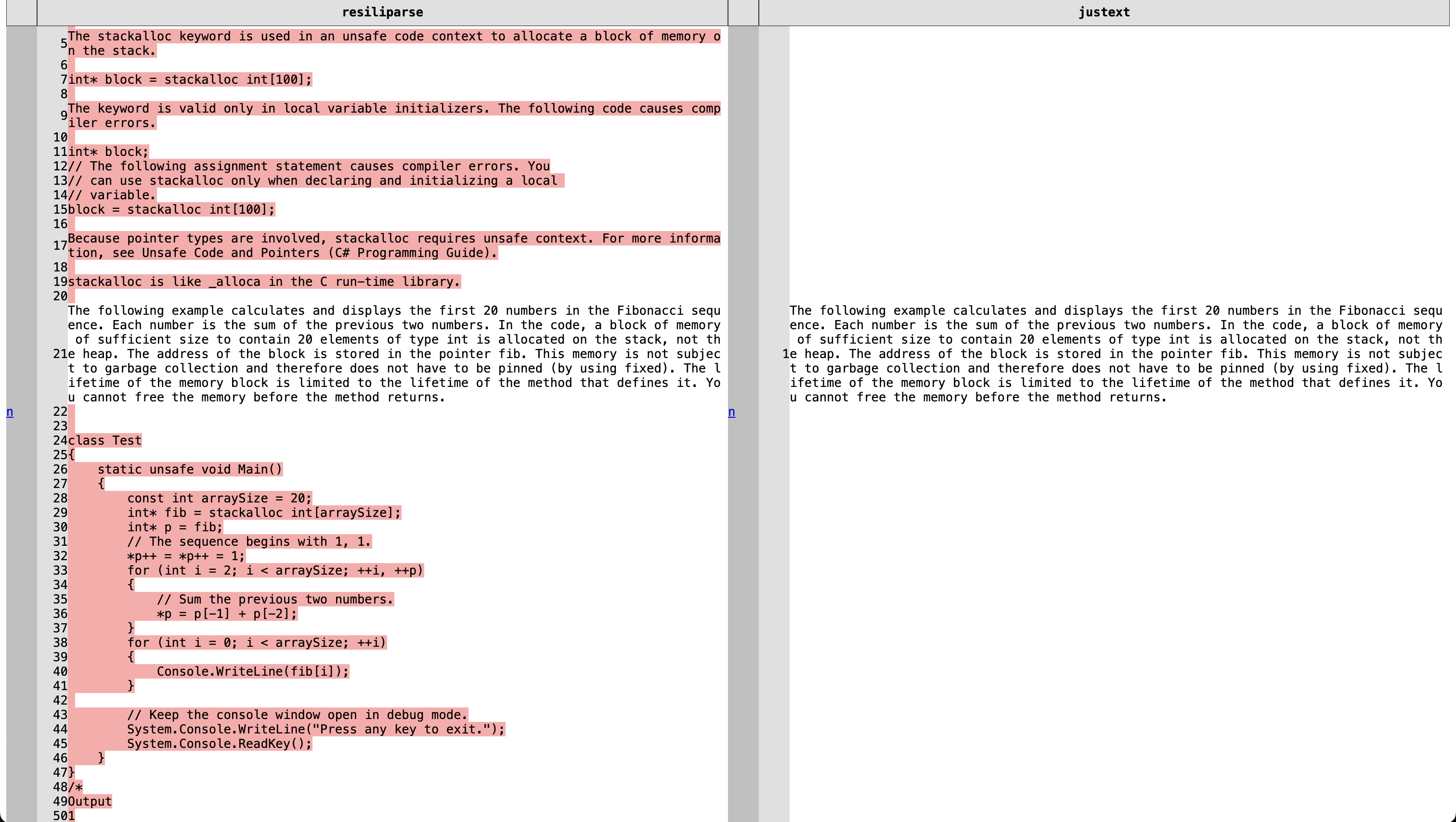}
    
    \vspace{6mm}
    
    \includegraphics[width=1.0\linewidth]{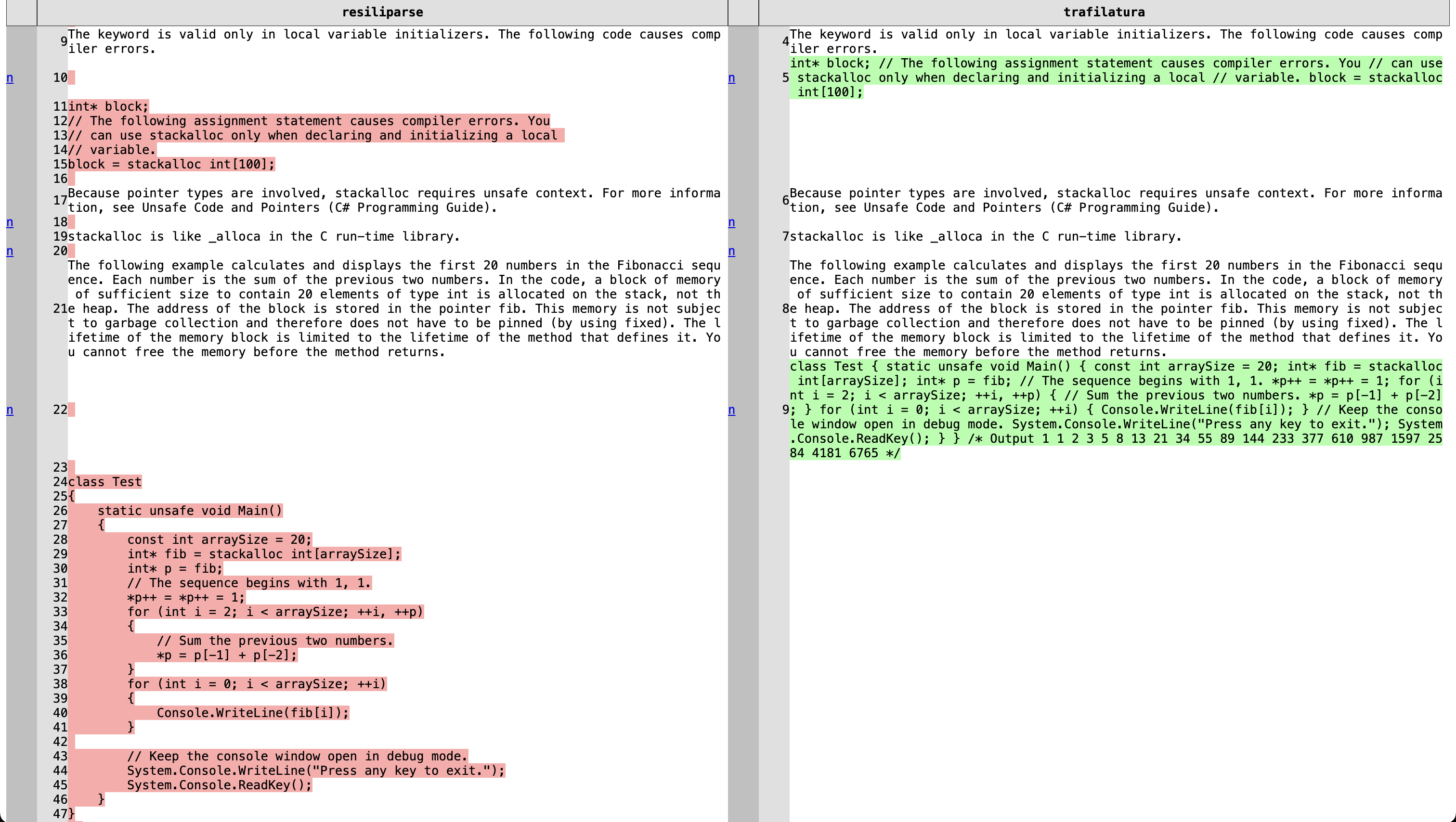}
    \caption{\textbf{Extraction comparison for a C\# tutorial page.} We use \texttt{difflib} to visualize pairwise comparisons between \resiliparse{} (left) and \justext{} (top) or \trafilatura{} (bottom). Note that for code blocks in this page, \justext{} removes them while \trafilatura{} collapses whitespace formatting.}
    \label{fig:code_example_1}
\end{figure}

\clearpage
\noindent\centering\small\textbf{Example 2:} \url{http://techbase.kde.org/index.php?title=Development/Tutorials/Plasma/Ruby/Using_widgets&diff=37860&oldid=37859}

\begin{figure}[h!]
    \centering
    \includegraphics[width=1.0\linewidth]{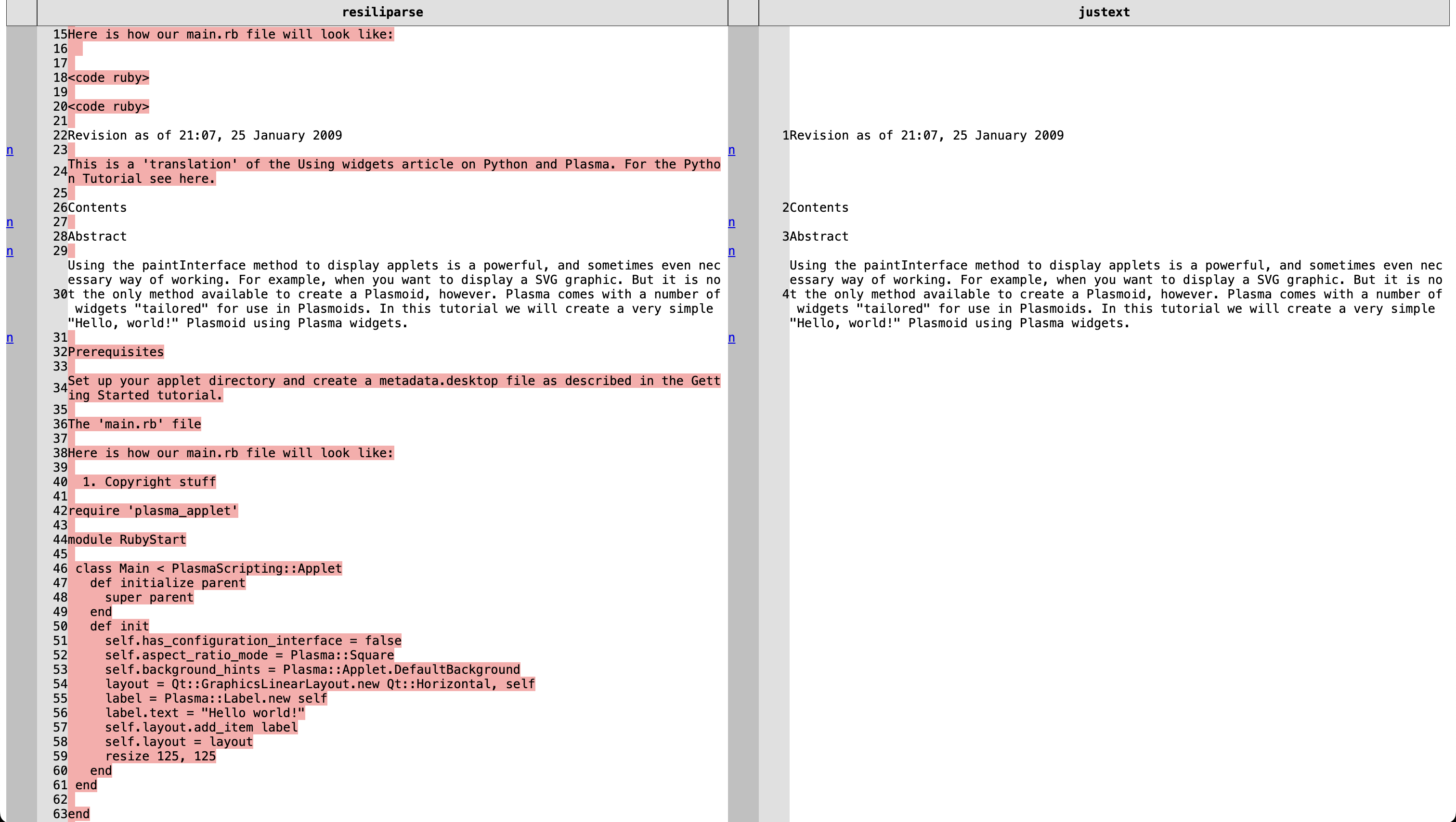}
    
    \vspace{6mm}
    
    \includegraphics[width=1.0\linewidth]{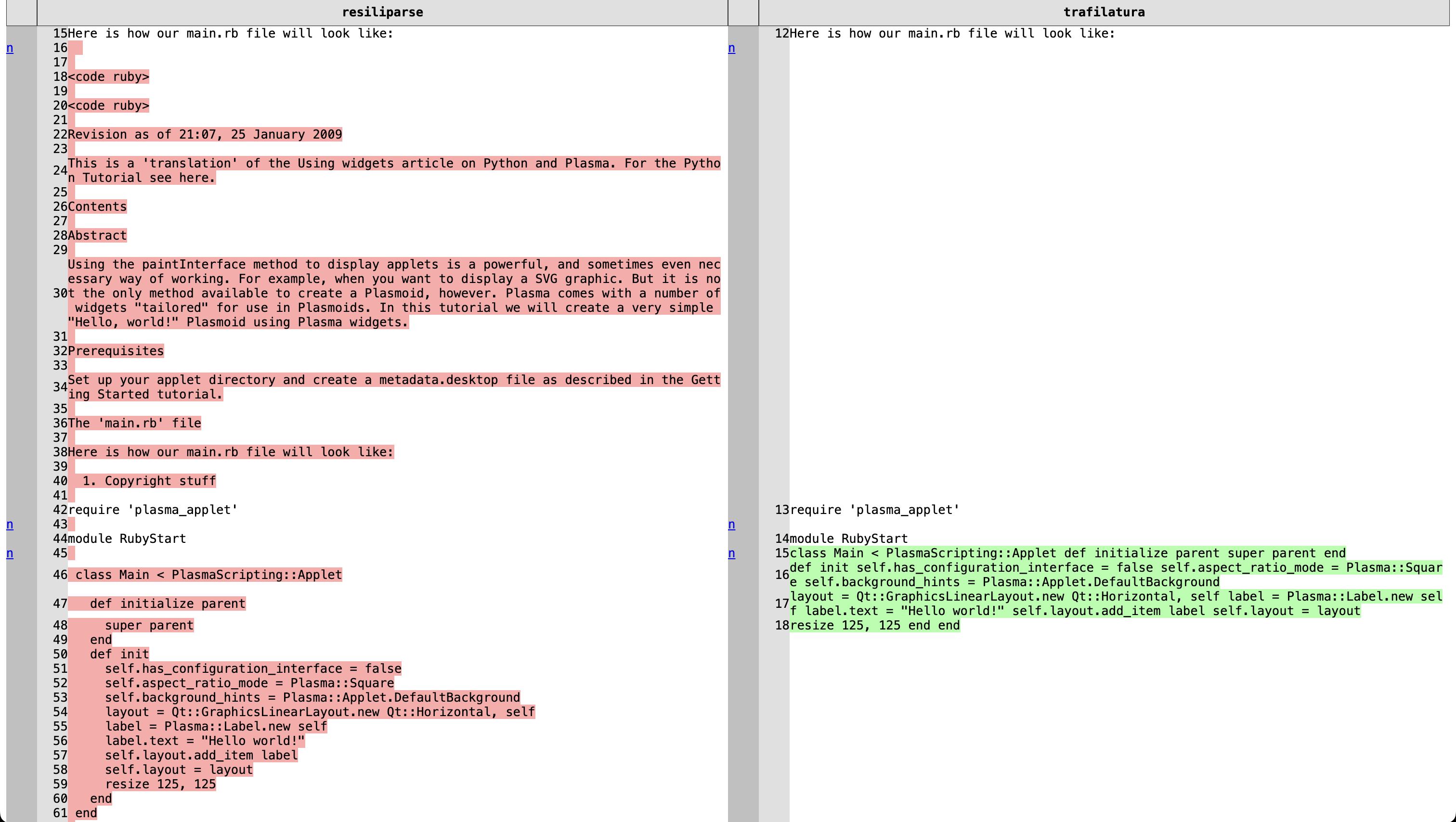}
    \caption{\textbf{Extraction comparison for a Ruby Plasma tutorial.} We use \texttt{difflib} to visualize pairwise comparisons between \resiliparse{} (left) and \justext{} (top) or \trafilatura{} (bottom). Note that for code blocks in this page, \justext{} removes them while \trafilatura{} collapses whitespace formatting.}
    \label{fig:code_example_2}
\end{figure}

\clearpage
\noindent\centering\small\textbf{Example 3:} \url{ http://www.advogato.org/person/knipknap/diary.html?start=82}

\begin{figure*}[h!]
    \centering
    \includegraphics[width=1.0\linewidth]{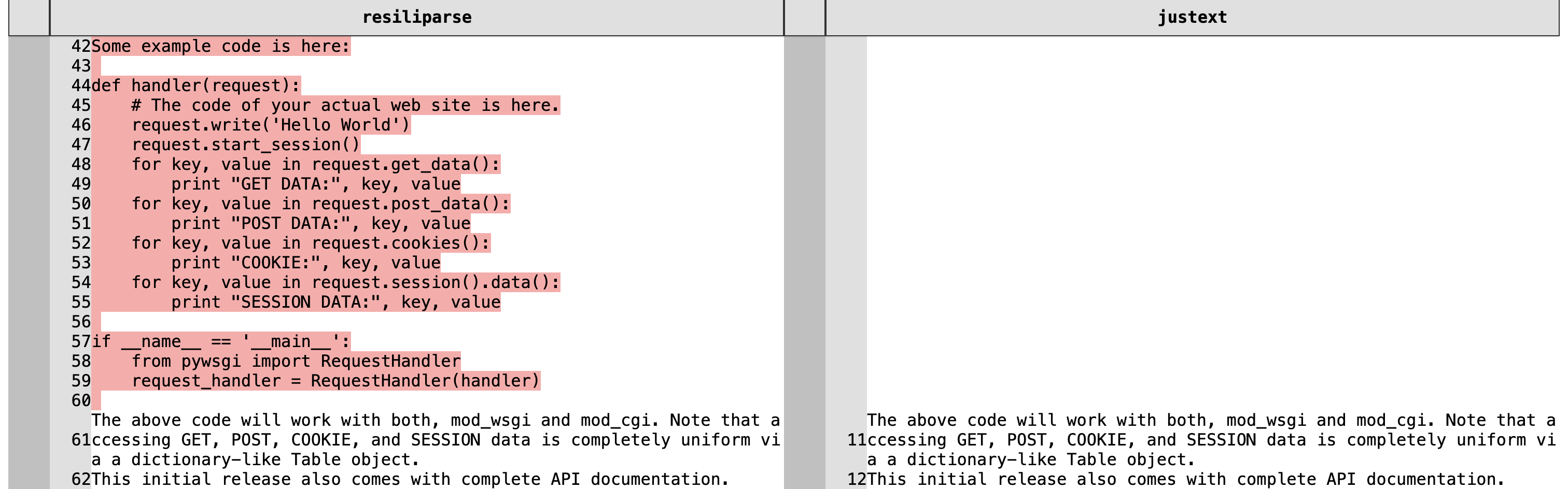}
    
    \vspace{6mm}
    
    \includegraphics[width=1.0\linewidth]{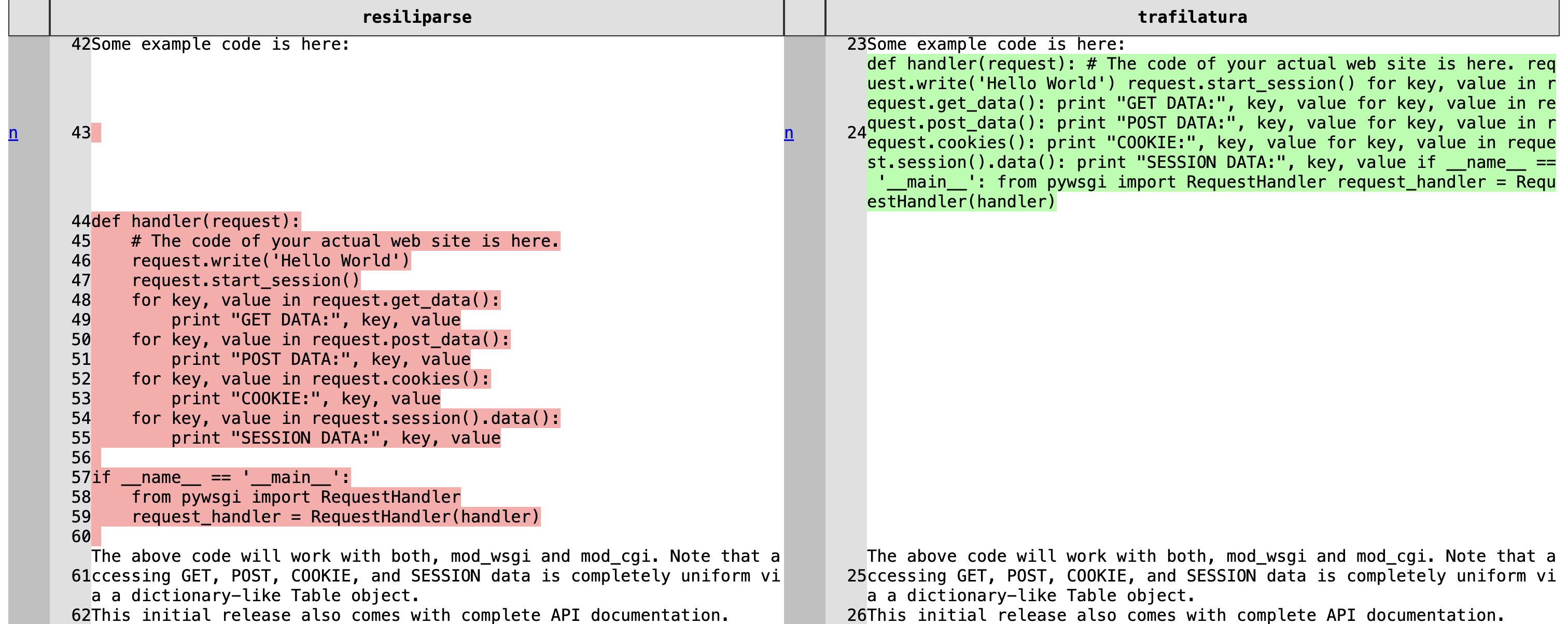}
    \caption{\textbf{Extraction comparison for release note blog post.} We use \texttt{difflib} to visualize pairwise comparisons between \resiliparse{} (left) and \justext{} (top) or \resiliparse{} (bottom). Note that for code blocks in this page, \justext{} removes them while \trafilatura{} collapses whitespace formatting.}
    \label{fig:code_example_3}
\end{figure*}

\normalsize \justifying \twocolumn

\clearpage

\subsection{Filtering for CC-Code}\label{app:cc_code_filtering}

We filter the same starting \xltrack{} pool corresponding to DCLM-RefinedWeb. We first retain only pages that contain \texttt{<pre>} elements, focusing on \texttt{<pre>} since it is a common way to incorporate code blocks while preserving whitespace formatting (e.g., indentation). However, \texttt{<pre>} is also used for other types of pre-formatted content such as lyrics, poems, or mailing threads. We therefore train a \fasttext{} classifier to better identify genuine code blocks.

\textbf{Content-based filtering for code blocks.} We inspect the 160 domains that contain the most pages with \texttt{<pre>} elements, labeling each domain as either code or non-code related (when unambiguous). We use this categorization to label all \texttt{<pre>} elements from these source pages as positive and negative. For positives, we also include any \texttt{<pre>} elements that contain a child \texttt{<code>} element. As with the classifier we used for CC-Tables, we train and use a model based on \textit{tokenized} HTML inputs. For code, however, we also ensemble this with an additional classifier that operates on an extracted version obtained using the  \texttt{get\_text()} function from \texttt{BeautifulSoup}\footnote{\url{https://pypi.org/project/beautifulsoup4/}} which keeps just the visible text. Based upon inspecting classifier scores, we averaged the scores from both and use  0.9 threshold. After this content-based filtering, we run our extractors and perform the same English filtering as is done by DCLM-Baseline but with a slightly looser threshold of 0.25 to account for the presence of code. We also use DCLM's \fasttext{} quality classifier to filter down to 55B tokens (to ensure sufficient size for no repeats in the cooldown setting).

\section{Training details}

For our training runs, we generally follow the fixed training recipes (i.e., architectures and hyperparameters) specified by DCLM~\citep{dclm} for their various competition scales and as implemented in \texttt{OpenLM}~\citep{open_lm}. We ran each \strack{} experiment on 4 nodes of 8xH100 GPUs which, as also indicated in Table 1 from \citet{dclm}, uses 240 total GPU hours. For \ltrack{} and \xltrack{} experiments, we use 16 nodes and incur costs of 3.7K and 7.3K GPU hours respectively.

A set of experiments that we ran outside of these standard training configurations is the alternative cooldowns in \Cref{tab:code_results} (top). For these, we still use the same number of concurrent GPUs as for \ltrack{} and \xltrack{} but the total cost now is 5.5K GPU hours due to the token count falling in between those scales. We also change the starting learning rate to resume from where it left off in the original DCLM-7B model's training run (close to 1e-3). 

Another set of experiments where training details differ are for our table experiments which make use of the Dataset Decomposition~\citep{pouransari2024dataset} technique to train longer-context models. We train for the same number of tokens as indicated by the scale name. However, the training runs incur an extra overhead of about 40\%. \strack{} models are trained for 4 length curriculum cycles while \ltrack{}  models are trained for 8. Both otherwise retain hyperparameters from the standard runs. For \trafilatura{} and \justext{}, we find it can help performance to use an intermediate checkpoint (saved after each cycle) but this does not meaningfully close the gap to \resiliparse{}.

\section{Evaluation details}

Given that we are evaluating pretrained language models, our test sets are separately curated English evaluations (as opposed to splits of our training data). Specifically, we use:

\begin{itemize}
    \item DCLM's~\citep{dclm} \lowvar{} evlauations include 22 different base model evaluations with varying test set sizes (please refer to \citet{dclm} for more details). 
    \item MMLU~\citep{hendrycks2020measuring} contains around 14K 4-way multiple choice questions covering 57 different subject categories
    \item For WikiTQ~\citep{pasupat-liang-2016-inferring}, we modify the implementation from \citet{torr} and evaluate each method on 100 instances $\times$ 7 serializations. 
    \item For HumanEval~\citep{Chen2021EvaluatingLL}, we use the implementation from \citet{bigcode-evaluation-harness} and evaluate each method on 164 examples $\times$ 5 languages. 
\end{itemize}

\newpage

\end{document}